\documentclass[10pt,twocolumn,letterpaper]{article}

\usepackage{pdfpages}

\usepackage{cvpr}
\usepackage{times}
\usepackage{epsfig}
\usepackage{graphicx}
\usepackage{amsmath}
\usepackage{amssymb} 
\usepackage{url}

\newcommand{\R}{\mathbb{R}}

\newcommand{\N}{\mathbb{N}}

\DeclareMathOperator*{\argmin}{arg\,min}

\newcommand{\shapeX}{\mathcal{X}}
\newcommand{\shapeY}{\mathcal{Y}}

\newcommand{\faces}{\mathcal{F}}
\newcommand{\SO}{\operatorname{SO}}
\newcommand{\SE}{\operatorname{SE}}

\newcommand{\perm}{\mathbb{P}}
\newcommand{\onevec}{\mathbf{1}}
\newcommand{\zerovec}{\mathbf{0}}

\makeatletter
\@tfor\next:=abcdefghijklmnopqrstuvwxyz\do{%
  \def\command@factory#1{%
    \expandafter\def\csname vec#1\endcsname{\mathbf{#1}}
  }
 \expandafter\command@factory\next
}

\@tfor\next:=ABCDEFGHIJKLMNOPQRSTUVWXYZ\do{%
  \def\command@factory#1{%
    \expandafter\def\csname mat#1\endcsname{\mathbf{#1}}
  }
 \expandafter\command@factory\next
}

\@tfor\next:=ABCDEFGHIJKLMNOPQRSTUVWXYZ\do{%
  \def\command@factory#1{%
    \expandafter\def\csname set#1\endcsname{\mathcal{#1}}
  }
 \expandafter\command@factory\next
}

\def\greekvectors#1{%
 \@for\next:=#1\do{%
    \def\X##1;{%
     \expandafter\def\csname mat##1\endcsname{\boldsymbol{\csname##1\endcsname}}
     }
   \expandafter\X\next;
  }
}

\greekvectors{alpha,beta,iota,gamma,lambda,nu,eta,Gamma,varsigma,Phi,Theta}

\makeatother

\usepackage{capt-of,etoolbox} %

\usepackage{nopageno} 

\usepackage{enumitem}  
\setitemize{noitemsep,topsep=0pt,parsep=0pt,partopsep=0pt}

\usepackage[pagebackref=true,breaklinks=true,letterpaper=true,colorlinks,bookmarks=false]{hyperref}

\cvprfinalcopy %

\def\cvprPaperID{8262} %

\ifcvprfinal\fi
\begin{document}

\title{MINA: Convex Mixed-Integer Programming for Non-Rigid Shape Alignment}

\author{Florian Bernard$\qquad$
Zeeshan Khan Suri$\qquad$
Christian Theobalt
\\
$\newline$
\\
MPI Informatics$\qquad$
Saarland Informatics Campus
}

\makeatletter
\let\@oldmaketitle\@maketitle%
\renewcommand{\@maketitle}{\@oldmaketitle%
  \myfigure{}\bigskip}%
\makeatother

\makeatletter
\newcommand{\settitle}{\@maketitle}
\makeatother

\newcommand\myfigure{%
  \centerline{
  \includegraphics[width=\linewidth]{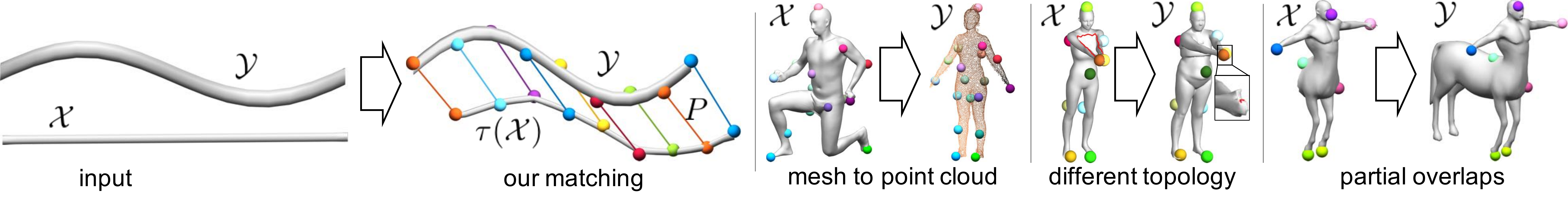}
  }
  \vspace{-1.5mm}
    \captionof{figure}{Left: MINA matches the straight cylinder $\shapeX$ non-rigidly to the curved cylinder $\shapeY$ (their orientations are arbitrary, but we visualise them with same orientation). It finds the \emph{globally optimal} correspondences $P \in \perm$ between a sparse set of points (coloured dots), as well as the non-rigid deformation $\tau$ of $\shapeX$ so that $\tau(\shapeX)$ aligns with $\shapeY$. MINA is highly flexible,~e.g.~it can match a mesh to a point cloud (middle left), match shapes with different topologies (middle right, where the hands in $\shapeX$ are \emph{not} touching, while the hands in $\shapeY$ are, see the geodesic paths between both hands shown as red lines), or deal with partial overlaps (right). (Best viewed on screen when zoomed in)}
    \label{fig:teaser}
  } 

\maketitle

\begin{abstract}
We present a convex mixed-integer programming formulation for non-rigid shape matching. To this end, we propose a novel shape deformation model based on an efficient low-dimensional discrete model, so that finding a globally optimal solution is tractable in (most) practical cases. Our approach combines several favourable properties: it is independent of the initialisation, it is much more efficient to solve to global optimality compared to analogous quadratic assignment problem formulations, and it is highly flexible in terms of the variants of matching problems it can handle. Experimentally we demonstrate that our approach outperforms existing methods for sparse shape matching, that it can be used for initialising dense shape matching methods, and we showcase its flexibility on several examples.
\end{abstract}

\section{Introduction}
Finding correspondences in geometric data is a long-standing problem in vision, graphics, and beyond. The applications range from the creation of statistical shape models, 3D reconstruction, 
 object tracking, or recognition, to more recent settings such as the alignment of geometric data to enable the training of deep learning models. In this work %
 we consider the problem of finding correspondences between two given shapes,  known as \emph{shape matching}.  

We assume that one shape is a geometrically transformed version of the other shape. With that, matching  shape $\shapeX$ to  shape $\shapeY$ can be phrased as finding a transformation
 $\tau$ (which belongs to a particular class $\Omega$ of transformations) such that the transformed shape $\tau(\shapeX)$ best aligns with  $\shapeY$. Formally, this can be written as the optimisation problem
 \begin{align}\label{eq:shapematching}
     \min_{\tau \in \Omega} \quad d(\tau(\shapeX), \shapeY)\,,
 \end{align}
 where $d(\cdot, \cdot)$ is a suitable metric that quantifies the discrepancy between both shapes. The particular shape matching setting depends on the choice of the metric $d(\cdot,\cdot)$ and the class of transformations $\Omega$. For example, \emph{rigid shape matching} refers to $\Omega = \SE(d)$, where $\SE(d)$ is the special Euclidean group in dimension $d$.
 In this work we study the \emph{non-rigid shape matching} problem, where $\Omega$ comprises non-rigid deformations (to be defined in Sec.~\ref{sec:nonrigidmodel}).
 
Although many previous works have addressed non-rigid shape matching, there are several open challenges:
    (i) due to the non-convex nature of Problem~\eqref{eq:shapematching} for virtually all relevant choices of $d(\cdot, \cdot)$ and $\Omega$, existing methods cannot guarantee to find global optima. Hence, these methods heavily depend on the initial choice of $\tau$.
    (ii) Oftentimes, non-rigid shape matching methods require that both shapes have the same representation (e.g.~meshes).
    (iii) Existing approaches have a limited flexibility in terms of the matching formulation that can be handled,~e.g.~they can only handle bijective matchings, or they cannot guarantee injectivity, they do not allow for additional constraints (e.g.~bounding the maximum distortion of a matching), or they cannot deal with shapes that have different topologies.
    (iv) Moreover, existing formulations that purely aim for preserving pairwise distances when finding a matching (see \emph{quadratic assignment problem} in Sec.~\ref{sec:relatedwork}) are not guaranteed to maintain the orientation of the surface.

\paragraph{Our contribution.} Our main idea is to formulate non-rigid shape matching in terms of a convex mixed-integer programming (MIP) problem, while addressing (i)-(iv). We summarise our main contributions as follows:
\begin{itemize}
    \item We propose a low-dimensional discrete model for non-rigid shape matching that is highly flexible as it allows to tackle a wide range of matching formulations. 
    \item Although solving MIP problems to global optimality has worst-case time complexity that is exponential in the number of integer variables,
    our proposed formulation only requires a small number of integer variables that is \emph{independent} of the shape resolution.
    \item Our formulation does not require an initialisation and it is oftentimes possible to (certifiably) find a globally optimal solution in practice.
\end{itemize}
 
\section{Related Work \& Background}\label{sec:relatedwork}
Due to the vast amount of literature related to shape matching and correspondence problems, it is beyond the scope of this paper to provide an exhaustive background of related work. A broad overview of the topic is for example presented in~\cite{van2011survey}. In the following we summarise works that we consider most relevant. 

\paragraph{Rigid shape matching.} Finding a rotation and translation that aligns two  shapes is known as \emph{rigid} shape matching.
The Procrustes problem~\cite{schonemann1966generalized} considers the setting when the correspondences between points on both shapes are known, which admits an efficient closed-form solution. However, rigid shape matching becomes significantly harder if the correspondences are unknown. Most commonly, this is addressed via local optimisation. A popular approach is  the Iterative Closest Point (ICP) algorithm~\cite{besl1992method}, which also comes in various variants, such as a probabilistic formulation~\cite{myronenko2010point}. These methods have in common that they do not guarantee to find a globally optimal solution and therefore their outcome is highly dependent on a good initialisation. Contrary to these local methods, for the rigid shape matching problem there are also global approaches,~e.g.~based on a semidefinite programming relaxation~\cite{maron2016point}, or on branch and bound algorithms~\cite{olsson2008branch,yang2013go}. 

A downside of these shape matching approaches is that they have the strong assumption that both shapes can be aligned based on a \emph{rigid-body} transformation. However, in practice this assumption is oftentimes violated, so that non-rigid shape matching approaches are more appropriate, which we will discuss next.

\paragraph{Functional maps.}
A popular paradigm for \emph{isometric} shape matching are 
functional maps (FM)~\cite{ovsjanikov2012functional,huang2017adjoint,ren2018continuous}, which define a framework for transferring a function from a source to a target shape.
Although FM were shown to be a powerful tool for isometric shape matching, they also have some shortcomings: they are sensitive to noise and suffer from symmetries, point-to-point maps obtained from FM are neither guaranteed to be smooth nor injective, and they are not suitable for severe non-isometries. {Rigid shape matching methods applied to spectral embeddings (obtained via FM) can also be used for isometric matching, such as done in PM-SDP~\cite{maron2016point}}. However, in this case the mentioned shortcomings also apply.

\begin{figure}
  \centering
    \includegraphics[width=0.8\linewidth]{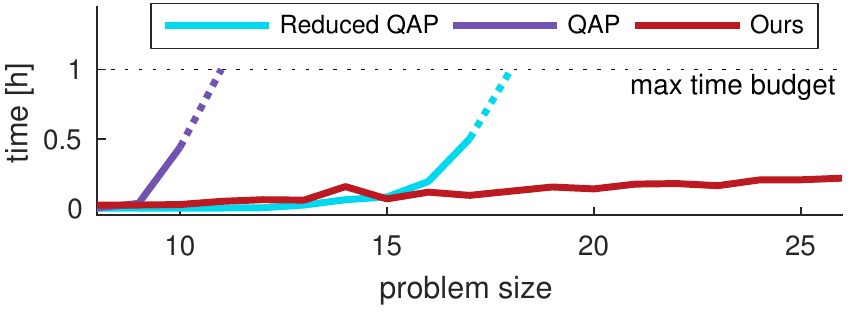}\hspace{6mm}
    \vspace{-1mm}
    \caption{Runtime comparison between our formulation, a QAP and a reduced QAP (cf.~search space reduction in the Supp. Mat.) when solved with MOSEK~\cite{mosek}.}
    \label{fig:qaptime} 
\end{figure}

\paragraph{Quadratic assignment problem.}
Another popular approach for non-rigid shape matching are formulations based on the quadratic assignment problem (QAP) (or graph matching)~\cite{Loiola:2ua4FrR7}, which aim for a non-rigid deformation with small distortion. In a discrete setting this can be phrased as matching vertices between two shapes in such a way that pairwise geodesic distances (or similar quantities) are (approximately) preserved by the vertex-to-vertex correspondences. The QAP is known to be NP-hard~\cite{rendl1994quadratic}, so that most solution approaches are based on heuristic approaches without formal guarantees, such as~e.g.~\cite{le2017alternating}. There are also more principled methods based on convex relaxations, including lifting-free~\cite{Zhou:2016ty,Dym:2017ue,bernard:2018} and lifting-based relaxations~\cite{Schellewald:2005up,swoboda2017b,kushinsky2019sinkhorn}. However, they do not guarantee to find a globally optimal solution of the \emph{original} non-convex problem as they rely on some kind of rounding procedure to obtain a binary solution. Globally optimal QAP solvers are  based on combinatorial search,~e.g.~via branch and bound~\cite{bazaraa1979exact}, and these methods scale exponentially in the number of variables.
Similarly as in the QAP, our method also takes the spatial context of matchings into account,
but we demonstrate that in practice our proposed formulation is significantly faster to solve,~cf.~Fig.~\ref{fig:qaptime}. {We believe that this is because our formulation has a special structure (based on our sparse deformation model) that can more efficiently be leveraged by combinatorial solvers.}

\paragraph{Global non-rigid matching.}
It was shown that certain matching problems can be solved globally optimal by finding shortest paths in a graph, or based on dynamic programming. These include matching 2D shapes (contours) to a 2D image~\cite{coughlan2000efficient,felzenszwalb2005representation,schoenemann2009combinatorial}, or matching a 2D contour to a 3D shape~\cite{lahner2016efficient}. As for example pointed out in~\cite{bernard2017combinatorial}, non-rigidly matching two objects in 3D is a significantly more difficult problem as it does not allow such a formulation. In~\cite{windheuser2011geometrically} the elastic matching of two 3D meshes is addressed based on a linear programming formulation. However, the formulation is sensitive to the mesh triangulation, requires a large number of binary variables, and {due to the non-tightness of relaxation it relies on sophisticated rounding techniques after which global optimality cannot be guaranteed anymore.}
In~\cite{chen2015robust} the authors propose a convex formulation for nonrigid registration that is solved via message passing. This approach requires an extrinsic term in order to disambiguate intrinsic symmetries, which in practice means that an initial alignment between both shapes is indispensable, thereby mitigating the advantages of a convex formulation.

\paragraph{Local non-rigid matching.} In a similar spirit, local refinement techniques also rely on a good matching initialisation. Such methods include~\cite{vestner2017product,vestner2017efficient} and~\cite{melzi2019zoomout}, where a given initial matching is gradually refined.
While ~\cite{vestner2017product} relies on a QAP formulation, in~\cite{melzi2019zoomout} a spectral method based on FM is used for a hierarchical upsampling. In Sec.~\ref{sec:densematching} we show that our method can be used as initialisation for such methods.

In~\cite{sumner2004deformation} the authors propose a non-rigid deformation model based on per-triangle affine transformations. Within this framework they also pose a correspondence problem, which, however, requires a good initial alignment between both shapes in order to make the optimisation problem well-posed.
Moreover, since the problem is non-convex, in general one only finds local optima. In our work we leverage a similar deformation model, but (i) we phrase the problem in a well-posed way without requiring an initial shape alignment, and (ii) we perform a global optimisation.

\paragraph{Learning-based matching.}
Shape matching has also been tackled with machine learning techniques,~e.g.~with random forests~\cite{rodola2014dense}, supervised deep functional maps~\cite{litany2017deep}, deep functional maps 
trained in self- or unsupervised settings~\cite{halimi2019unsupervised,roufosse2018unsupervised}, or using PointNet~\cite{qi2017pointnet} for learning point cloud correspondences~\cite{groueix2019unsupervised}. Undeniably, machine learning %
has the potential to address many open challenges in shape matching,~e.g.~for learning appropriate shape representations. %
In the past it was demonstrated that combinatorial shape matching benefits from learned deep features~\cite{bernard2017combinatorial}, and reversely, that embedding combinatorial optimisation solvers into neural networks (``differentiable programming'') opens up new possibilities for tackling a range of interesting matching problems~\cite{mena2018learning}.
We believe that in the future our  method may also be amenable to utilise such synergies, and therefore consider it to be orthogonal to learning-based methods.

{\paragraph{Convex mixed-integer programming.}
Mixed-integer programming refers to optimisation problems that involve both continuous and discrete variables. Their advantage is that they are extremely flexible and allow to model a wide range of complex problems. For example, they can be used to discretise difficult non-convex problems, such as formulations that impose rotation matrix constraints, or for phrasing matching problems with binary variables. 
However, the downside is that  MIP problems have a search space that has exponential size in the number of discrete variables, so that in general it is very hard to solve large problems to global optimality.
\emph{Convex mixed-integer programming} refers to a subclass of MIP problems that are \emph{convex} for fixed integer variables. A major advantage is that for this class of problems there exist efficient branch and bound solvers that globally optimise such problems.
 Albeit the fact that these solvers have a worst-case runtime that is  exponential in the number of integer variables, in this work we  demonstrate that solving non-rigid shape matching using a convex MIP reformulation is tractable in (most) practical scenarios.
}

\section{Non-Rigid Shape Matching}\label{sec:nonrigidmodel}

First, we summarise our notation. For an integer $i \in \N$ we define $[i] := \{1,\ldots, n\}$. For a matrix $X \in \R^{p \times q}$ and the index set $\mathcal{I} \subseteq [p]$ we use $X_{\mathcal{I}} \in \R^{|\mathcal{I}| \times q}$ to denote the $|\mathcal{I}|$ rows of $X$ selected by $\mathcal{I}$. $\onevec_n$ and $\matI_n$ denote the $n$-dimensional vector of all ones and the $n$-dimensional identity matrix, $\|\cdot\|$ denotes the Frobenius norm, and matrix and vector inequalities are understood element-wise.

Let $\shapeX$ %
and
$\shapeY$ %
be triangular surface meshes that are discretisations of Riemannian $2$-manifolds embedded in 3D space. Note that later in Sec.~\ref{sec:flex} we will also address the case when $\shapeY$ is a point cloud.
Our aim is to find a non-rigid deformation $\tau$ that transforms shape $\shapeX$ to $\tau(\shapeX)$, so that it aligns well with shape $\shapeY$, cf.~Problem~\eqref{eq:shapematching}.
For notational convenience we use $\shapeX \in \R^{n_{\shapeX} \times 3}$ and $\shapeY \in \R^{n_{\shapeY} \times 3}$ to refer to the matrices containing the $n_{\shapeX}$ and $n_{\shapeY}$ 3D vertex positions of shapes $\shapeX$ and $\shapeY$, respectively. Moreover, let $\faces^{\shapeX} \in [n_{\shapeX}]^{f_{\shapeX} \times 3}$ 
be  a matrix that encodes the triangular faces of $\shapeX$,
where $f_\shapeX$ 
is the number of triangles.

\subsection{Non-Rigid Deformation Model}
We model the non-rigid deformation of $\shapeX$ by applying an affine transformation to each triangle. In conjunction with suitable mesh consistency constraints, the individual per-triangle affine deformations globally constitute a non-rigid deformation. Although related deformation models have been introduced before~\cite{sorkine2007rigid,sumner2004deformation,bernard2017combinatorial,sumner2007embedded}, they have not been used for a global optimisation of non-rigid shape matching.

\paragraph{Affine per-triangle transformations.} For the $i$-th vertex $\shapeX_i$ we define the non-rigid deformation $\tau_p$ in terms of its adjacent triangle $p$ as{}
\begin{align}\label{eq:pertridef}
  \tau_p(\shapeX_i) := (\shapeX_i - c_p) Q_p + c_p + t_p \,,
\end{align}
where %
$c_p \in \R^{1{\times}3}$ is the centroid of the $p$-th triangle in the undeformed shape $\shapeX$, $Q_p \in \R^{3{\times}3}$ is a linear transformation, and $t_p \in \R^{1{\times}3}$ is a translation. As such, we first centre a given vertex, apply a linear transformation, undo the centring, and eventually translate it to its global position.

\begin{figure}
\centering{
    \includegraphics[width=0.9\linewidth]{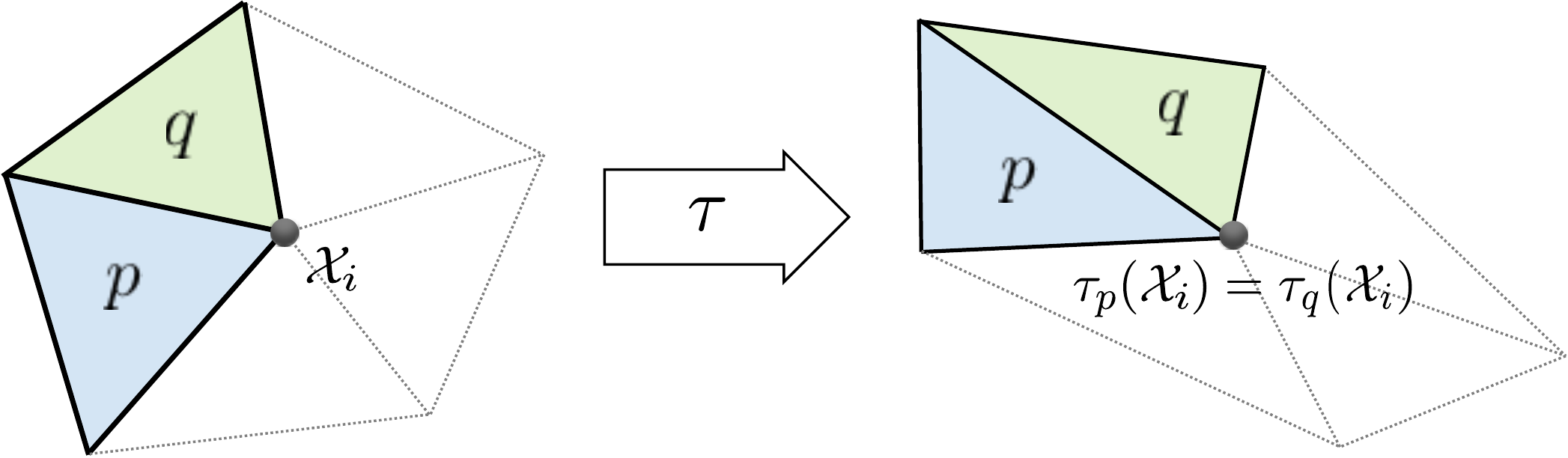}
}
    \caption{Illustration of the deformation $\tau$, %
     where each triangle undergoes an affine transformation. The consistency constraint imposes that the transformed $\shapeX_i$ is the same, no matter whether it is transformed by $\tau_p$ or $\tau_q$ for $p \in \mathcal{N}_i$ and $q \in \mathcal{N}_i$.}
    \label{fig:consistency} 
\end{figure}

\paragraph{Mesh consistency constraints.} In order to ensure a consistent mesh deformation, we impose the constraints
\begin{align}\label{eq:meshcons}
  \tau_p(\shapeX_i) = \tau_q(\shapeX_i) \quad \forall ~i \in [n_\shapeX],~p \in \mathcal{N}_i,~q \in \mathcal{N}_i \,,
\end{align}
where $\mathcal{N}_i \subset [f_\shapeX]$ is the set of all triangles in $\shapeX$ that are adjacent to vertex $i$,~cf.~Fig.~\ref{fig:consistency}. The purpose of the constraints~\eqref{eq:meshcons} is to enforce that a given vertex $\shapeX_i$ is transformed to the same place, no matter which transformation of its adjacent triangles is applied, and thereby ensuring that the triangle topology is preserved by the deformation.

\subsection{Mixed-Integer Non-Rigid Shape Alignment} \label{sec:mina}
\paragraph{Low-dimensional correspondence model.} Our non-rigid deformation $\tau$ is indirectly defined by a low-dimensional discrete model.
To this end,
subsets of the shape vertices are used as \emph{control points}, which are represented by the matrices $\shapeX_{\mathcal{I}} \in \R^{u \times 3}$ and $\shapeY_{\mathcal{J}} \in \R^{v \times 3}$. Here, $u:= |\mathcal{I}|$ and $v:=|\mathcal{J}|$ denote the total number of control points for each shape, and $\mathcal{I} \subseteq [n_{\shapeX}]$ and $\mathcal{J} \subseteq [n_{\shapeY}]$ denote the index sets that select the control points from the original shapes. A similar approach has been pursued in~\cite{sumner2007embedded} for the interactive \emph{manipulation} of shapes, where, however it is assumed that for each control point of $\shapeX_{\mathcal{I}}$ the corresponding control point of $\shapeY_{\mathcal{J}}$ is already known.  In contrast, we are interested in the much more difficult shape \emph{matching} problem, where the correspondence between control points is \emph{unknown}, and, moreover, there may not even exist an \emph{exact} counterpart in $\shapeY_{\mathcal{J}}$ for each point in $\shapeX_{\mathcal{I}}$. 

\begin{figure}[t!]
\centerline{
\scalebox{0.8}{
  \begin{tabular}{@{}c@{}}{\vspace{-1cm}$\shapeX$}\end{tabular}
  \begin{tabular}{@{}c@{}}{\includegraphics[width=0.33\linewidth]{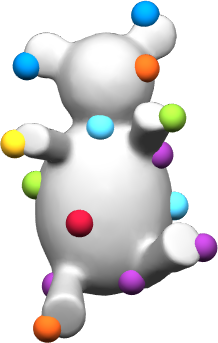}}\end{tabular}
  ~~~~~~
  \begin{tabular}{@{}c@{}}{\includegraphics[width=0.4\linewidth]{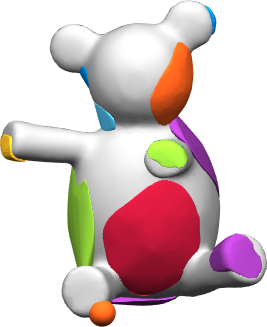}}\end{tabular}
  \hspace{-5mm}
  \begin{tabular}{@{}c@{}}{\vspace{-0.9cm}$\shapeY$}\end{tabular}
}
}
    \caption{Illustration of control points of $\shapeX$ (left) that are matched to convex polyhedra of $\shapeY$ (right). Colours indicate  correspondences between control points and convex polyhedra. 
    }
    \label{fig:ptToCvxPoly} 
\end{figure}

\paragraph{Convex polyhedral surface approximation.}
We propose to address the issue that there may not exist exact counterparts between control points as follows:
rather than matching control points of $\shapeX_{\mathcal{I}}$ directly to control points of $\shapeY_{\mathcal{J}}$, 
we match the points of $\shapeX_{\mathcal{I}}$ to \emph{convex polyhedra} that locally approximate the surface of $\shapeY$, see~Fig.~\ref{fig:ptToCvxPoly}. To this end, we associate a convex polyhedron with each control point of $\shapeY_{\mathcal{J}}$, which we represent using the matrix $Z_j \in \R^{d_j{\times}3}$ for $j \in [v]$. Here, each row of $Z_j$ contains one of the $d_j \in \N$ vertices (corner points) of the $j$-th polyhedron on $\shapeY$. As such, any point that lies inside the convex polyhedron can be specified as a convex combination of rows of $Z_j$,~i.e.~$\theta^T Z_j$, where the $d_j$-dimensional vector $\theta$ satisfies the \emph{convex combination constraints} $\theta \geq \zerovec_{d_j}$ and $\theta^T \onevec_{d_j} = 1$. 
We note that this point-to-polyhedron matching is a strict generalisation of point-to-point matching, since the latter is achieved for $d_j {=} 1$.
Using this formulation allows 
 to find a matching between $\shapeX_{\mathcal{I}}$ and $\shapeY_{\mathcal{J}}$ even when there exists only an approximate counterpart between the control points on both shapes. {For details how we obtain the  polyhedra see the Supp. Mat.}

\paragraph{Correspondence term.} We tackle the non-rigid shape matching problem by establishing correspondences between the control points $\shapeX_{\mathcal{I}}$ and the convex polyhedra  of $\shapeY$, while at the same time ensuring that the resulting non-rigid deformation $\tau$ is ``regular''.
For now, let us assume that $u \leq v$ and that each control point of $\shapeX_{\mathcal{I}}$ is matched to one of the convex polyhedra of $\shapeY$. Moreover, we also allow that more than one control point of $\shapeX_{\mathcal{I}}$ can be matched to the same convex polyhedron of $\shapeY$. 
We model these \emph{matching constraints} using the matrix $P \in \perm_{uv}$, where we define
\begin{align}
  \perm_{uv} := \{ P \in \{0,1\}^{u{\times}v}~:~ P \onevec_v = \onevec_u \}\,. %
\end{align}
An element $P_{ij}{=}1$ means that the $i$-th control point of $\shapeX_{\mathcal{I}}$ is matched to the $j$-th convex polyhedron of $\shapeY$.
{Later, in Sec.~\ref{sec:outliers}, we will also present more general formulations
that allow to also match shapes when some control points on $\shapeX$ do not have a counterpart on $\shapeY$.}

Since there are $u$ control points of $\shapeX_{\mathcal{I}}$, where each of them is matched to one of the $v$ convex polyhedra on $\shapeY$, 
for each $i{\in}[u],j{\in}[v]$ we introduce a convex combination weight vector $\alpha_{ij} {\in }\R^{1{\times}d_j}$. Here, $i$ is the index of the control point of $\shapeX_{\mathcal{I}}$ and $j$ is the index of the convex polyhedron of $\shapeY$. By defining $d{:=}\sum_{j=1}^v d_j$, $Z {:=} [Z_1^T, \ldots, Z_v^T]^T \in \R^{d {\times}3}$, as well as the matrix of convex combination weights
\begin{align}
  \matalpha := [\alpha_{ij}]_{i\in[u],j\in[v]} \in \R^{u \times d}\,,
\end{align}
 we model our \emph{correspondence term} as
\begin{align} \label{eq:corrterm}
  f_{\text{corr}}(\tau, P) := \lambda_\text{c} \, \|  \tau(\shapeX_{\mathcal{I}}) - \matalpha Z \|\,. %
\end{align}
In addition we impose $\matalpha {\geq} \zerovec$, $\matalpha \onevec_{d} {=} \onevec_u$, and $\alpha_{ij} \onevec_{d_j} {\leq} P_{ij}$ for $i{\in}[u],j{\in}[v]$. As such, we can effectively enforce the convex combination and matching constraints using \emph{linear equalities}. With that, the $(u{\times}3)$-dimensional matrix $\matalpha Z$ contains points that lie inside the convex polyhedra, where each of its $u$ rows  correspond to the respective row of the transformed control point matrix $\tau(\shapeX_{\mathcal{I}}) \in \R^{u{\times}3}$. 

Moreover, to avoid that multiple control points are assigned to the same \emph{vertex} of a convex polyhedron, we impose the ``soft-injectivity'' constraints $\onevec_u^T \matalpha {\leq} \onevec_d^T$. {The soft-injectivity constraint enforces that  the sum of weights in each column of $\boldsymbol{\alpha}$ is at most one. As such, if a (single) element in a column is exactly one, only this control point is assigned to the respective vertex of the convex polyhedron. If elements in a column of $\boldsymbol{\alpha}$ are strictly smaller than one, all respective control points are assigned to \emph{non-extreme points} of the polyhedron, thereby preventing that multiple control points are matched to the same \emph{vertex} of a polyhedron.} 

We use the notation $(\matalpha,P) \in \Gamma$ to refer to the four constraints introduced in this paragraph.
In overall, the correspondence term has the purpose to minimise the discrepancy between the control points of the transformed shape $\tau(\shapeX_{\mathcal{I}})$ and their corresponding convex polyhedra of $\shapeY$. 

\paragraph{Deformation regularisers.} 
For regularising the deformation $\tau$ we decompose each linear transformation $Q_p$ in~\eqref{eq:pertridef} into the sum of a rotation matrix $R \in \SO(3)$ and a (small) general linear part $T_p \in \R^{3{\times}3}$, so that $\tau_p$ in~\eqref{eq:pertridef} now becomes
\begin{align}
    \tau_p(\shapeX_i) := (\shapeX_i - c_p) (R+T_p) + c_p + t_p \,.
\end{align}
The purpose of using the additive factorisation $Q_p = R + T_p$ (with $\|T_p\|$ small) is to ensure that the global shape deformation $\tau$  (approximately) preserves the morphology of $\shapeX$.
{This has a similar effect as the as-rigid-as-possible (ARAP) model~\cite{sorkine2007rigid}, but requires only a single rotation matrix compared to $f_{\shapeX}$ rotation matrices as used in ARAP.}
In order to keep the linear part $T_p$ small, we impose the \emph{rigidity loss} as
\begin{align}
  f_{\text{rigid}}(\tau) := \lambda_\text{r} \, \big\| \,[T_1, \ldots, T_{f_\shapeX}] \,\big\|\,.
\end{align}

Moreover, for achieving a locally smooth deformation, we introduce the \emph{smoothness loss} 
\begin{align}\label{eq:smoothness}
  f_{\text{smth}}(\tau) := \lambda_\text{s} \, \big\| \,[\omega_1\Delta_{1}, \ldots, \omega_{|\mathcal{E}|}\Delta_{|\mathcal{E}|}] \,\big\|\,,
\end{align}
where $\mathcal{E} \subset [f_\shapeX]^2$ denotes the set of all neighbouring triangle pairs in $\shapeX$, and $\omega_e$ is a scalar weight. For $e = (p,q) \in  \mathcal{E}$, so that triangles $p$ and $q$ are neighbours, we define the $e$-th smoothness residual as
\begin{align}
  \Delta_e &= \tau_p(c_q) - \tau_q(c_q) 
           = \tau_p(c_q) - (c_q + t_q)  \,.
\end{align}
The purpose of the residual $\Delta_e \in \R^{1{\times}3}$ is to quantify the difference between transforming the triangle centroid $c_q$ using the transformation $\tau_q$ of the same triangle, and using the transformation $\tau_p$ defined for its neighbour triangle $p$.

\paragraph{Optimisation problem.} Based on the introduced terms and constraints our mixed-integer non-rigid alignment (\emph{MINA}) formulation reads
\begin{align}\label{eq:cvxmip}
&\argmin_{\substack{P,\matalpha, R, \{T_p\},\{t_p\}}} ~~  && f_{\text{corr}}(\tau,P) + f_{\text{rigid}}(\tau) + f_{\text{smth}}(\tau)  \\
& ~~~~~\text{s.t.} &&  \tau_p(\shapeX_i) = \tau_q(\shapeX_i)\,, \nonumber \\
&&& P \in \perm_{uv}\,, \nonumber \\
&&& (\matalpha,P) \in \Gamma\,, \nonumber \\
&&& R \in \SO(3)\,. \nonumber 
\end{align}
We assume that all weights $\lambda_\bullet \geq 0$.
The mesh consistency and  the $\Gamma$ constraints are affine in the variables $P, R, \matalpha, T_p$ and $t_p$, and all objective function terms $f_{\bullet}$ are compositions of affine transformations with the Frobenius norm, so that they are convex. However, due to the binary constraints imposed upon $P$, and the non-convex quadratic equality constraints $R \in \SO(3)$, the overall problem is non-convex.

\begin{figure} 
       \centerline{ {\includegraphics[width=0.4\linewidth]{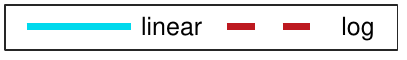}} }
       \vspace{-1mm} 
      \centerline{ 
        {\includegraphics[width=0.32\linewidth]{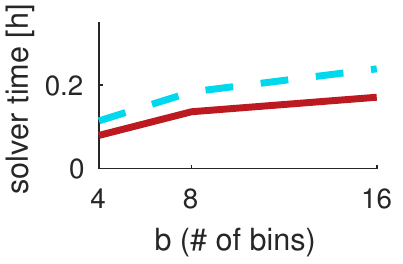}} \hfil
        {\includegraphics[width=0.32\linewidth]{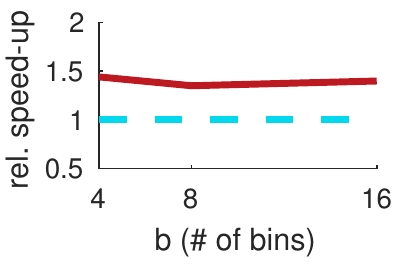}} \hfil
        {\includegraphics[width=0.32\linewidth]{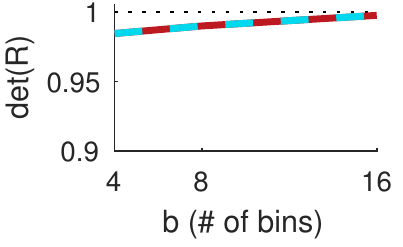}}
      }
      \vspace{-1mm}
    \caption{Runtime, relative speed-up~w.r.t.~\emph{linear}, and determinants of $R$ for \emph{linear} and \emph{logarithmic} encodings of the $\SO(3)$ discretisation.}
    \label{fig:ablationSOdiscretisation} 
\end{figure}

\paragraph{Convex mixed-integer formulation.} 
To transform Problem~\eqref{eq:cvxmip} into a convex MIP problem, we use a piece-wise linear approximation of the $\SO(3)$ constraint based on binary variables, see the Supp. Mat. and~\cite{dai2017global}. %
To keep the number of binary variables small, we use an efficient Gray encoding for the piece-wise linear approximation,~cf.~\cite{vielma2011modeling}, so that the number of binary variables is \emph{logarithmic} in the number of discretisation bins $b$. {The main idea here is to utilise a more efficient representation that requires fewer binary variables and thus admits a more efficient optimisation.}
In particular, this results in $6{\cdot}\lceil\log_2(b)\rceil$ binary variables, in contrast to $6{\cdot}b$ binary variables for a naive linear encoding.
In Fig.~\ref{fig:ablationSOdiscretisation} we compare our used logarithmic encoding with a linear one, where it can be seen that the logarithmic one requires less computation time, and that the determinant of the resulting matrix is already very close to $1$ for $b{=}4$.

\newcommand{\tabfig}[1]{\begin{tabular}{@{}c@{}} #1 \end{tabular}}

\begin{figure*}
\centerline{
\scalebox{0.67}{
    \tabfig{
    \includegraphics[height=2.5cm]{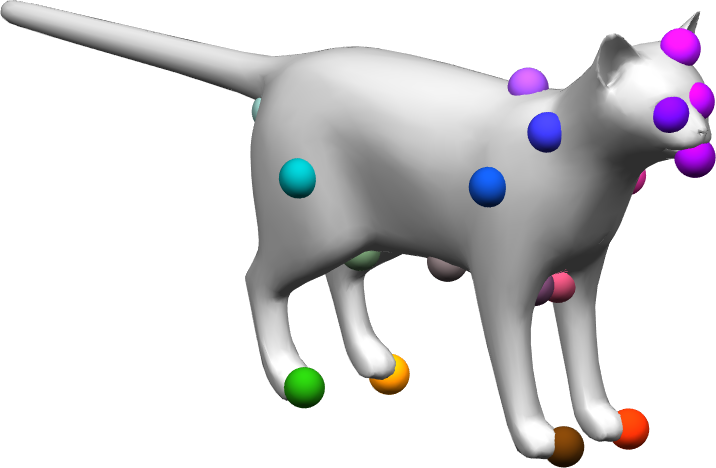}\\
    \includegraphics[height=2.3cm]{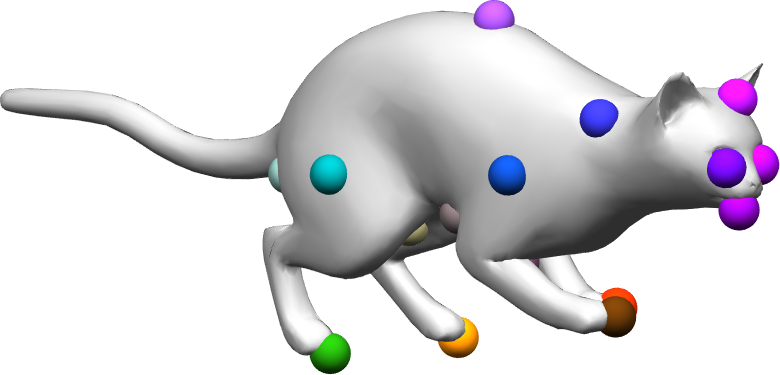}
    }
    \tabfig{
    \includegraphics[height=3cm]{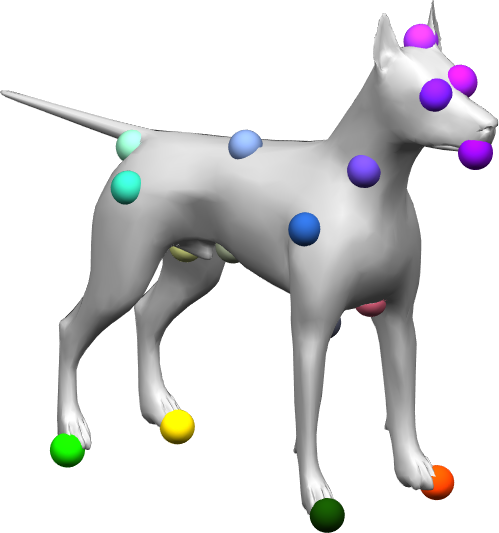}\\
    \includegraphics[height=2cm]{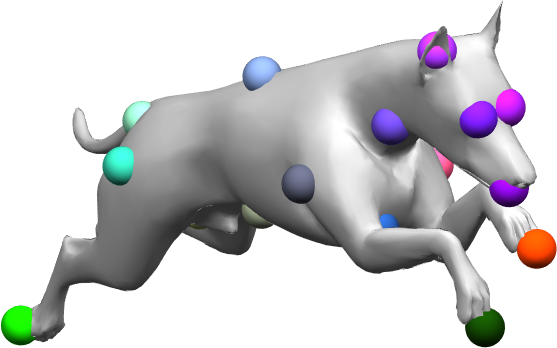}
    }
    \tabfig{\includegraphics[height=2.8cm]{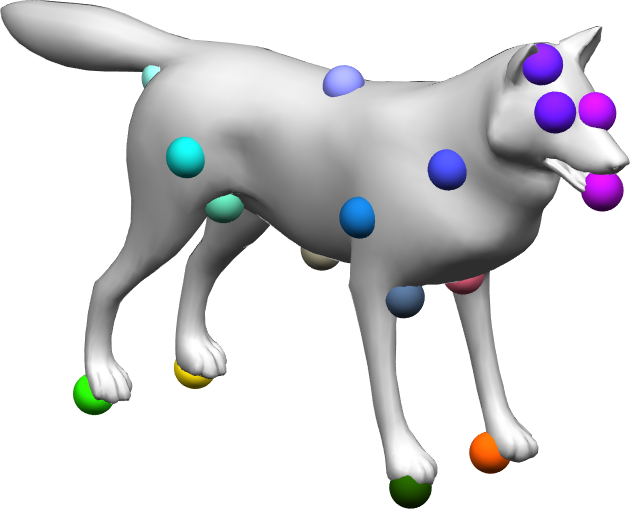}\\
    \includegraphics[height=2.8cm]{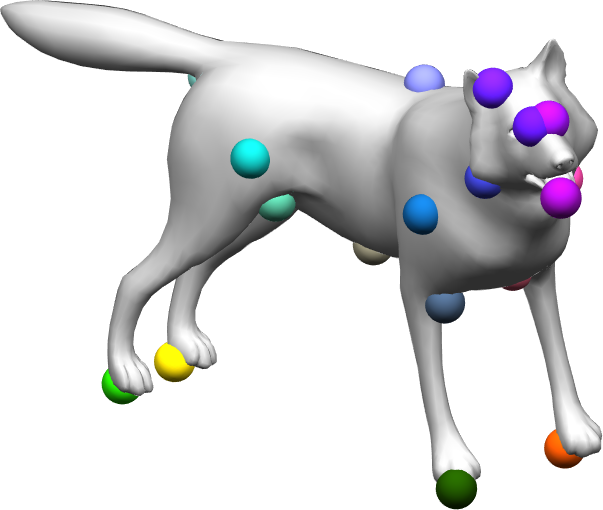}}
    ~~~~
    \tabfig{
    \includegraphics[height=3.5cm]{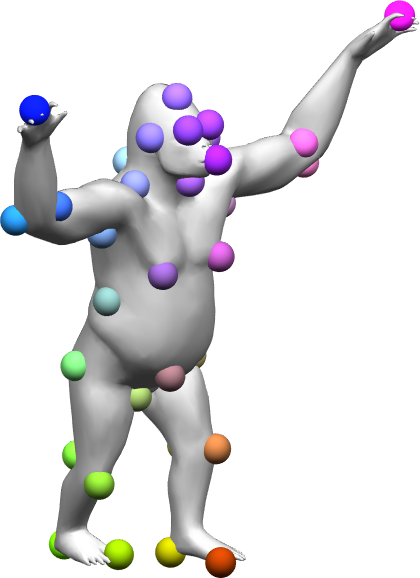}\\
    \includegraphics[height=2.5cm]{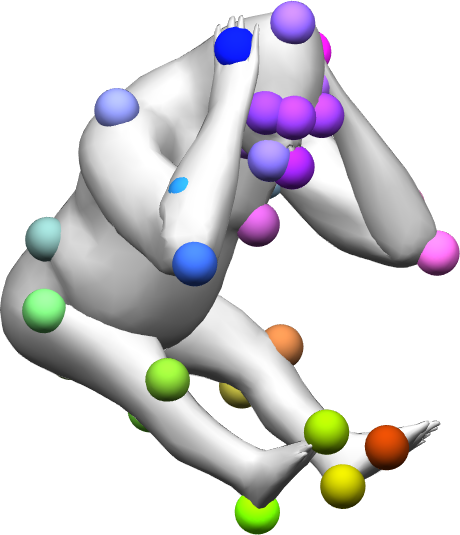}}
    \tabfig{\includegraphics[height=2.7cm]{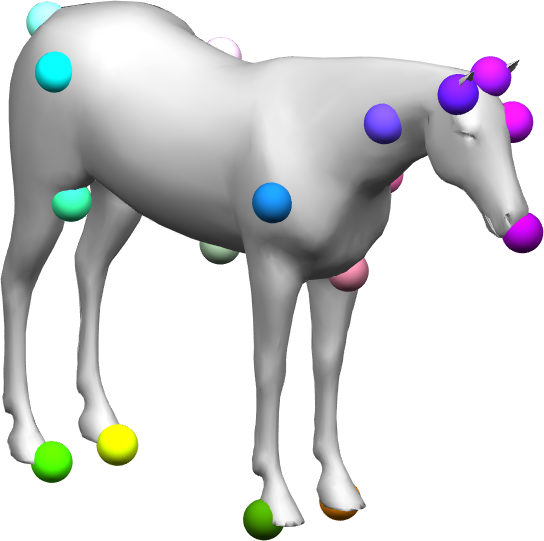}\\
    \includegraphics[height=2.6cm]{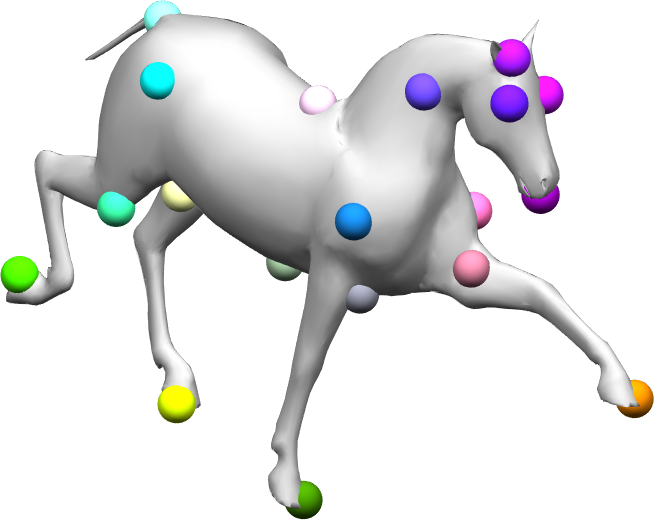}}
    \tabfig{\includegraphics[height=3cm]{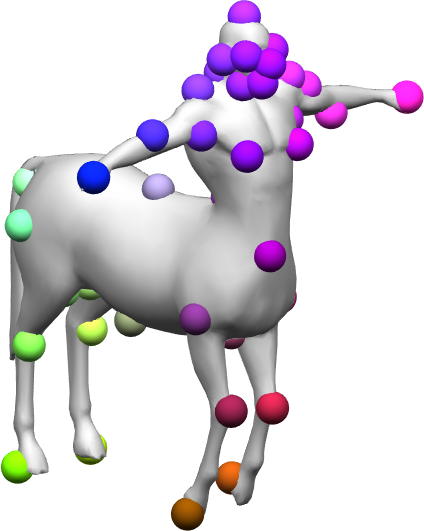}\\
    \includegraphics[height=3.0cm]{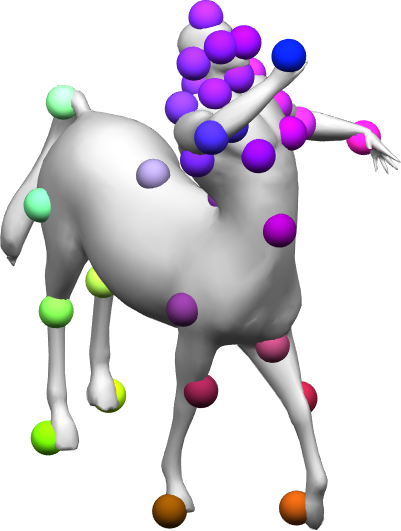}}
    \tabfig{\includegraphics[height=3cm]{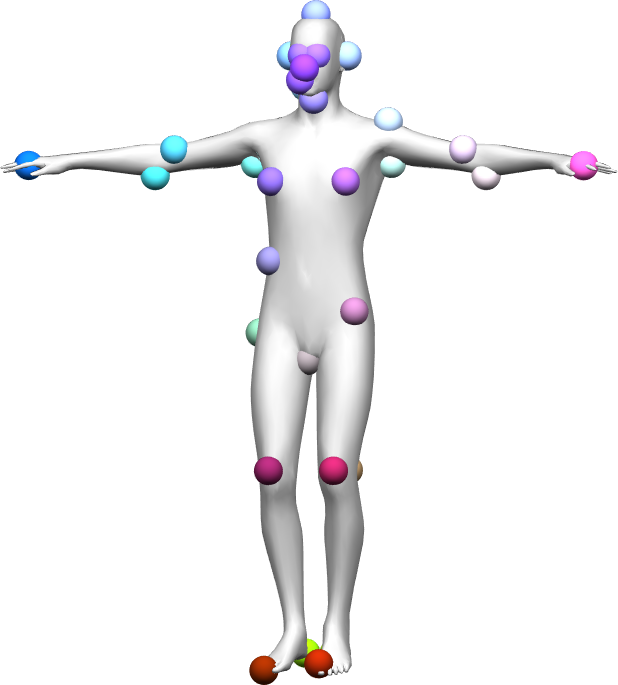}\\
    \includegraphics[height=3.0cm]{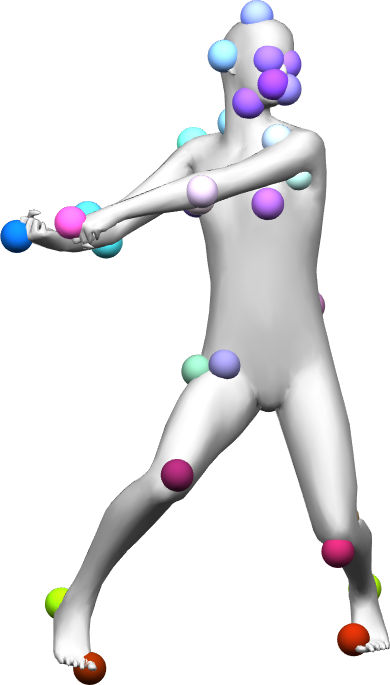}}
    \tabfig{\includegraphics[height=3cm]{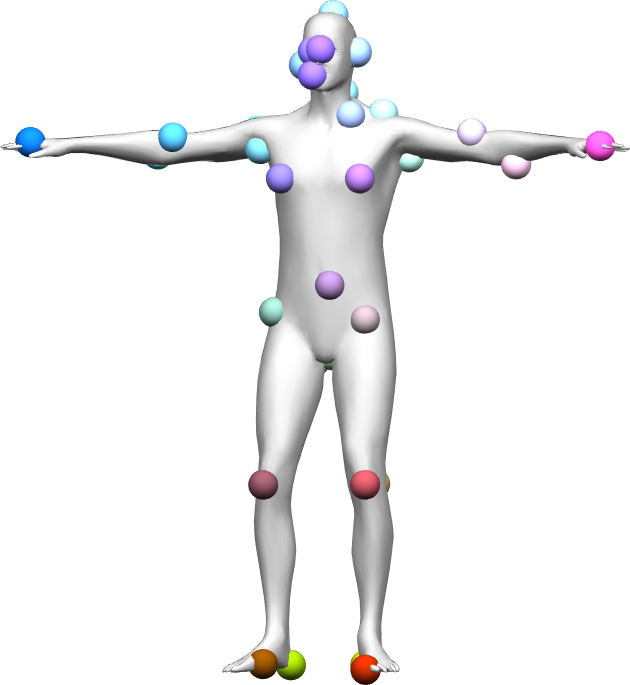}
    \vspace{3mm}\\
    \includegraphics[height=2.5cm]{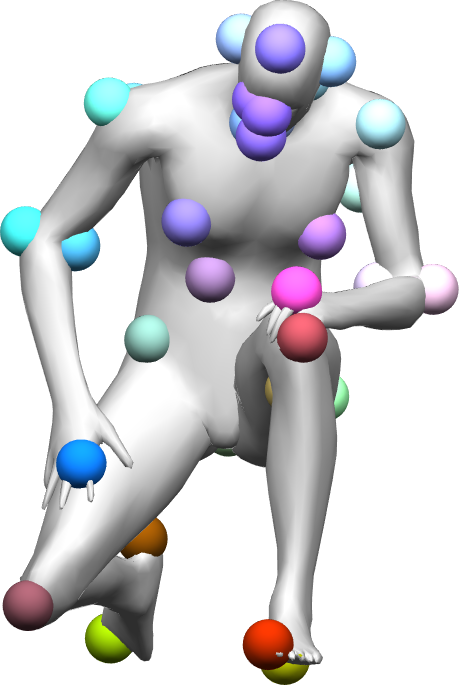}}
  }
}
    \caption{Correspondences obtained from our method for several shape matching instances from the TOSCA dataset~\cite{Bronstein:2008:NGN:1462123}. Correspondences between $\shapeX$ in the top row and $\shapeY$ in the bottom row are indicated by dots with corresponding colours.}
    \label{fig:qualitativeResultsTosca} 
\end{figure*}

\newcommand{\toscaPckWidth}{2.7cm}
\begin{figure}
       \centerline{ {\includegraphics[width=\linewidth]{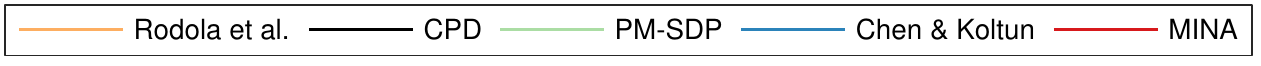}} }
      \centerline{ 
        {\includegraphics[width=\toscaPckWidth]{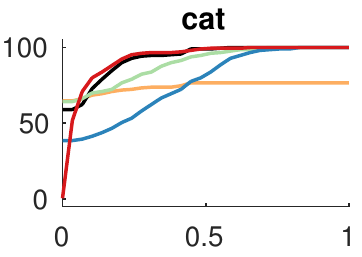}} \hfil
        {\includegraphics[width=\toscaPckWidth]{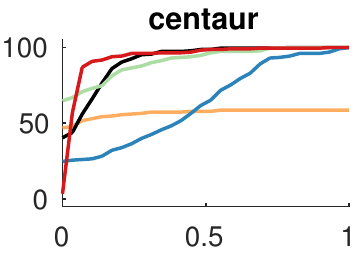}} \hfil
        {\includegraphics[width=\toscaPckWidth]{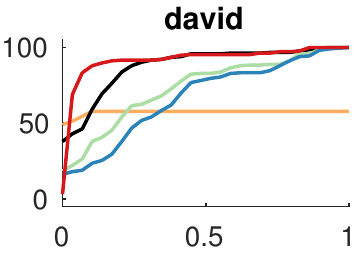}} 
      } 
      \vspace{1mm}
      \centerline{ 
        {\includegraphics[width=\toscaPckWidth]{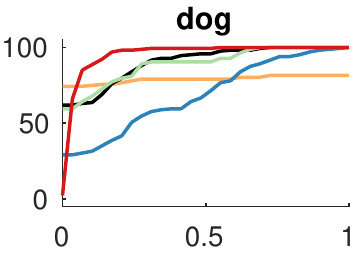}} \hfil
        {\includegraphics[width=\toscaPckWidth]{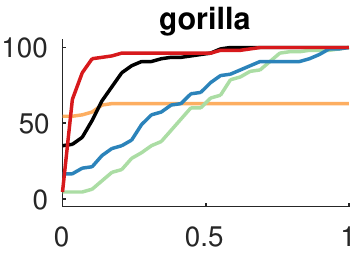}} \hfil
        {\includegraphics[width=\toscaPckWidth]{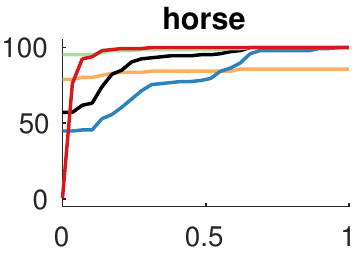}} 
      }
      \vspace{1mm}
      \centerline{ 
        {\includegraphics[width=\toscaPckWidth]{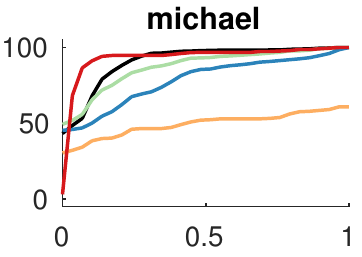}} \hfil
        {\includegraphics[width=\toscaPckWidth]{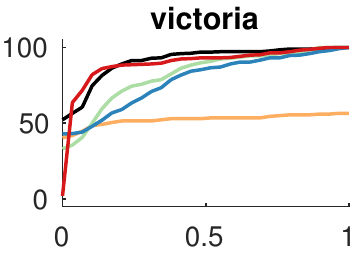}} \hfil
        {\includegraphics[width=\toscaPckWidth]{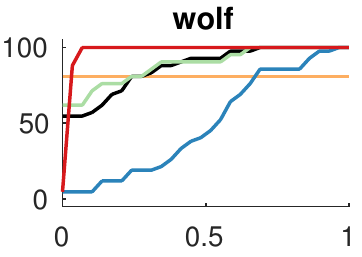}}
      }
    \caption{Each plot summarises the percentage of correct matchings for different TOSCA shape classes. The horizontal axis shows the geodesic error threshold, and the vertical axis shows the percentage of matches that are smaller than or equal to this error.}
    \label{fig:quantitativeResults} 
\end{figure}

\section{Experiments}

In this section we present an experimental evaluation of our proposed MINA approach. To this end, we compare it to other sparse correspondence methods, we analyse the gaps to global optimality,
we demonstrate that MINA can be used as initialisation for dense shape matching,
 and we showcase its flexibility on several exemplary settings.
We provide additional implementation details in the Supp. Mat..

\subsection{Sparse Shape Matching}\label{sec:sparseMatching}
In this section we compare our method with other approaches
that perform a sparse matching between a pair of shapes. In particular, we consider the convex matching method PM-SDP~\cite{maron2016point} in a rigid setting, the sparse game-theoretic approach by Rodola et al.~\cite{rodola2012game}, the coherent point drift (CPD) algorithm~\cite{myronenko2010point} {(randomly initialised)}, and the convex relaxation by Chen \& Koltun~\cite{chen2015robust}. As such, we cover a wide range of shape matching paradigms, including convex relaxations for rigid (PM-SDP) and non-rigid (Chen \& Koltun) shape matching, a local non-rigid method (CPD), and a sparse method that considers a quadratic assignment problem formulation (Rodola et al.). In this set of experiments we use the sparse points from~\cite{Kim11} for matching pairs of shapes from the TOSCA dataset~\cite{Bronstein:2008:NGN:1462123}. Hence, we directly match control points on $\shapeX$ to control points on $\shapeY$ when using our MINA method (i.e.~$d_j {=} 1$ for $j {\in} [v]$).

In Fig.~\ref{fig:qualitativeResultsTosca} we show  correspondences obtained from our method for various shape matching pairs.  In Fig.~\ref{fig:quantitativeResults} we show quantitative results, where we summarise the percentage of correct matches (relative to the number of given control points) for each shape class in the TOSCA dataset. It can be seen that our MINA method generally outperforms the other sparse matching approaches. {The lower scores for smaller geodesic thresholds arise due to our sparse modelling, since matchings can only be as accurate as the sparse control points allow for.} Since the method by Rodola et al.~\cite{rodola2012game} does not match all of the given points, the respective curves do not reach 100\%. {Moreover, the performance of PM-SDP indicates that a rigid matching setting is too restrictive.}
{Additional results can be found in the Supp. Mat.}

\subsection{Global Optimality Analysis} 
\begin{figure}[h]
       \centerline{ {\includegraphics[width=.4\linewidth]{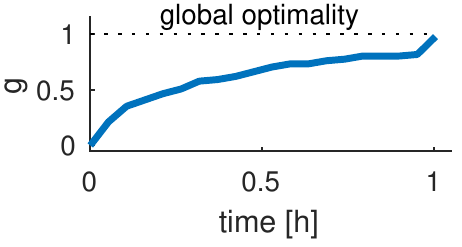}} ~~~\hfil
       {\includegraphics[width=.4\linewidth]{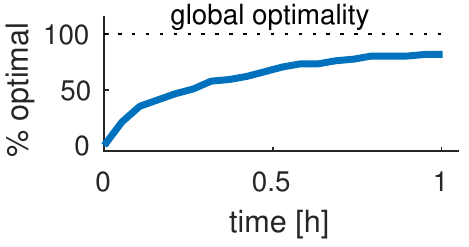}}
       }
       \vspace{-1mm}
    \caption{Optimality gaps (left) and proportion of instances solved to global optimality (right) dependent on the solver runtime.}
    \label{fig:relativeGapStats} 
\end{figure}

Here, we analyse the gaps to global optimality dependent on the processing time $t$ for the TOSCA shape matching instances in Sec.~\ref{sec:sparseMatching}. To this end, we define
\begin{align}
  g(t) = \frac{1}{N}\sum_{i: t_i \leq t} (1-\sigma^{\text{rel}}_i)\,,
\end{align}
where $N$ is the total number of shape matching pairs ($N{=}71$ for TOSCA), $t_i$  denotes the total solver time for the $i$-th shape matching problem, and $\sigma^{\text{rel}}_i$ is the \emph{relative gap} of the $i$-th problem that is defined as
$\sigma^{\text{rel}}_i {=} \frac{| \overline{f} - \underline{f} |}{\max(\delta, |\overline{f}|)}$ (see~\cite{mosek}).
Here, $\overline{f}$ and $\underline{f}$ are the upper and lower bounds
of the objective value  
of the MIP formulation of Problem~\eqref{eq:cvxmip}, respectively, and $\delta$ is a small number. In Fig.~\ref{fig:relativeGapStats} (left) it can be seen that after $1$h (our time budget for the MIP solver) the value of $g$ reaches $0.98$,~i.e.~on average the solutions are close to being globally optimal{~(a value of $1$ means that all instances are solved to global optimality). After $1$h, for $82\%$ of the cases we certify global optimality, see Fig.~\ref{fig:relativeGapStats} (right).}

\newcommand{\pigSpaceAfter}{0.4cm}
\newcommand{\pigSpaceBefore}{0.4cm}
\newcommand{\dogSpaceAfter}{0.4cm}
\newcommand{\dogSpaceBefore}{0.3cm}
\newcommand{\teddySpaceAfter}{0.3cm}
\newcommand{\teddySpaceBefore}{0.2cm}
\newcommand{\glassesSpaceAfter}{0.4cm}
\newcommand{\glassesSpaceBefore}{0.1cm}
\begin{figure*}
\centerline{
\scalebox{0.875}{
  \tabfig{
    \vspace{.5cm}
    \rotatebox{90}{\hspace{-1mm}Reference}~~\includegraphics[height=1.4cm]{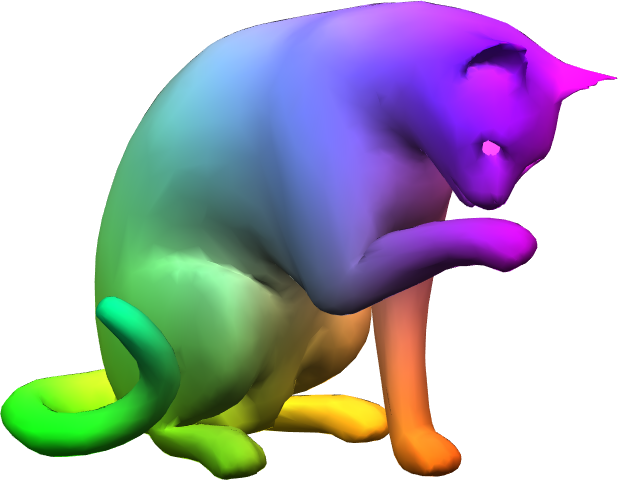}\vspace{.5cm}\\
    \rotatebox{90}{\hspace{7mm}random}~~\includegraphics[height=2.5cm]{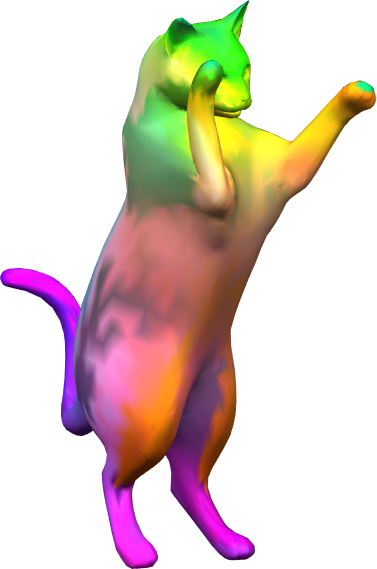}\\
    \rotatebox{90}{\hspace{6mm}PM-SDP}~~\includegraphics[height=2.5cm]{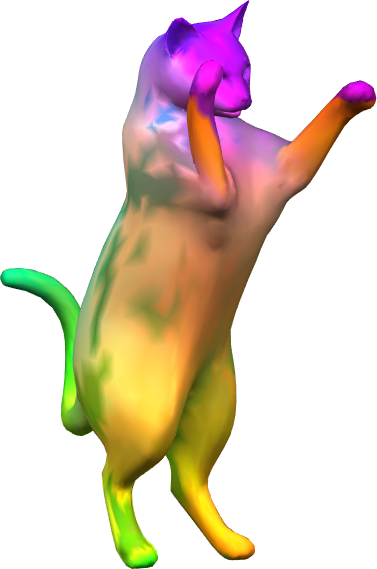}\\
    \rotatebox{90}{\hspace{3mm}Rodola~et~al.}~~\includegraphics[height=2.5cm]{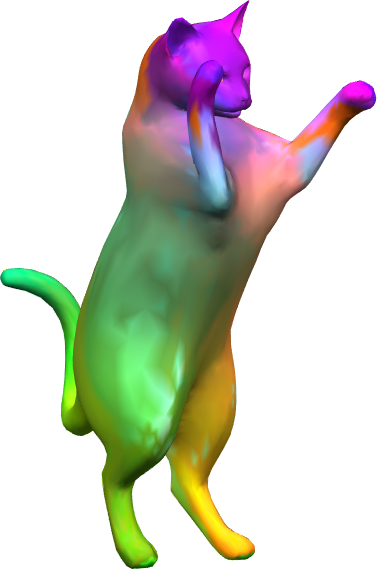}\\
    \rotatebox{90}{\hspace{7mm}MINA}~~\includegraphics[height=2.5cm]{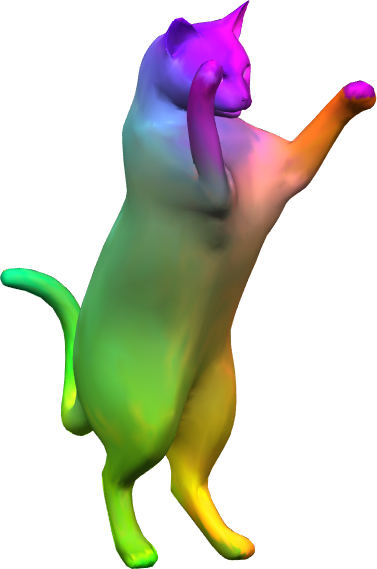}\\
  }
  \tabfig{
    \vspace{.2cm}\includegraphics[height=2cm]{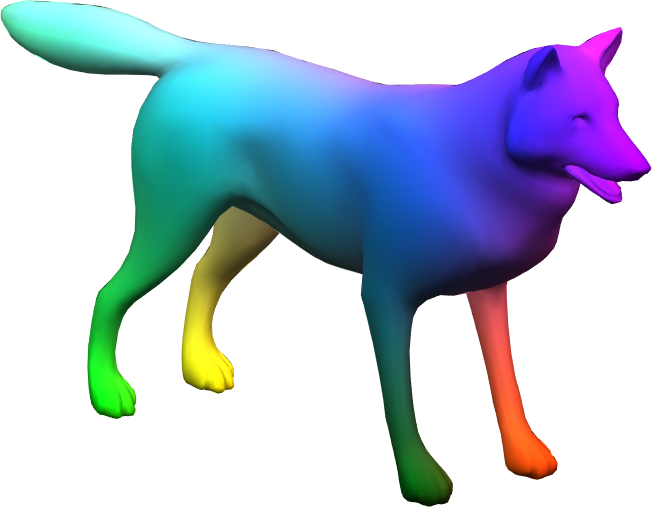}\\
    \includegraphics[height=2.5cm]{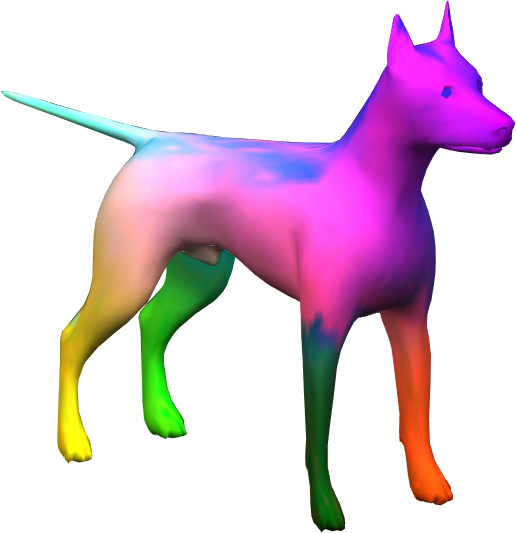}\\
    \includegraphics[height=2.5cm]{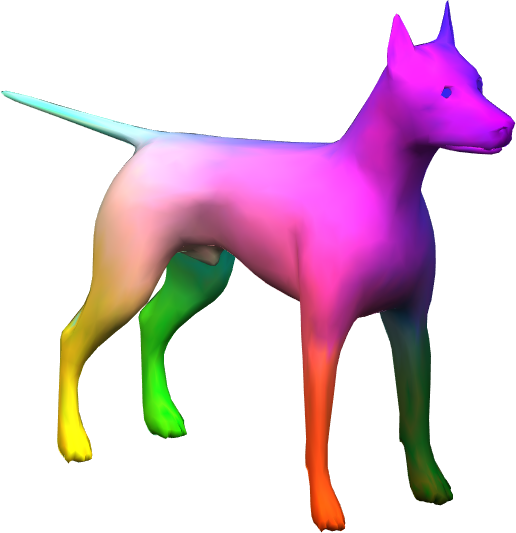}\\
    \includegraphics[height=2.5cm]{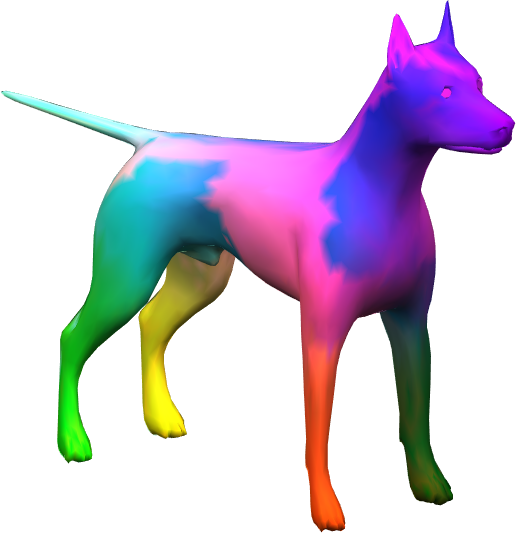}\\
    \includegraphics[height=2.5cm]{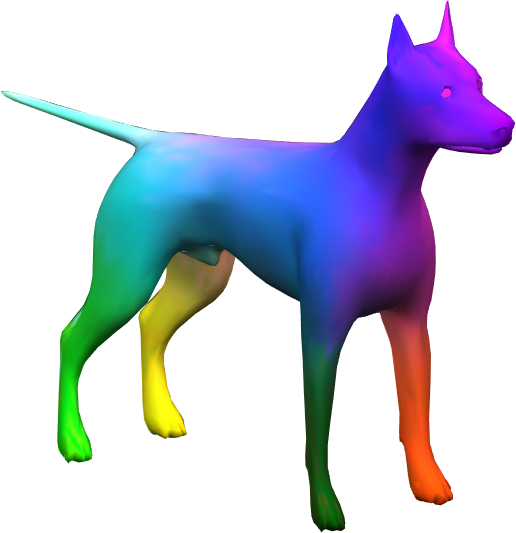}\vspace{.1cm}\\
  }
  \tabfig{
    \vspace{0cm}\includegraphics[height=2.5cm]{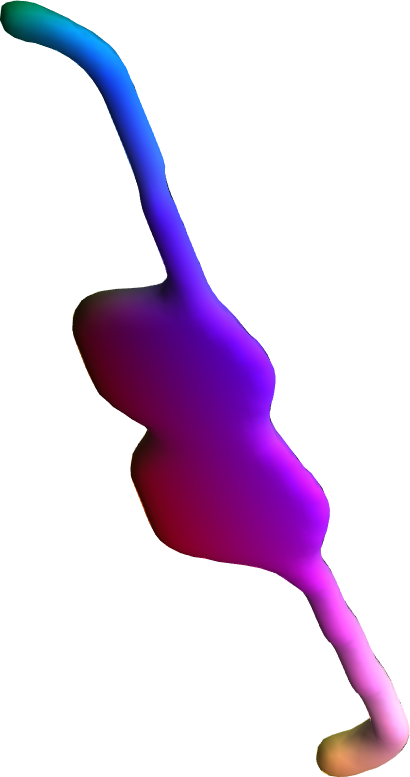}\vspace{0cm}\\
    \vspace{\glassesSpaceBefore}\includegraphics[height=2cm]{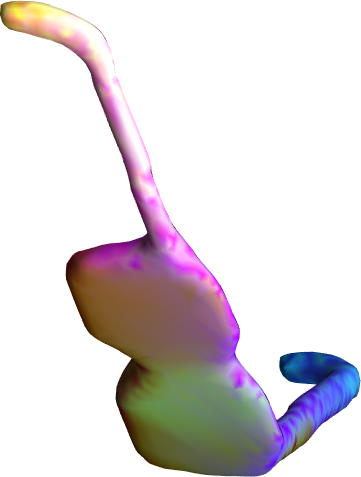}\vspace{\glassesSpaceAfter}\\
    \vspace{\glassesSpaceBefore}\includegraphics[height=2cm]{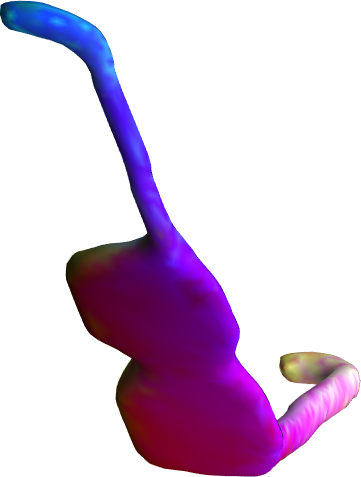}\vspace{\glassesSpaceAfter}\\
    \vspace{\glassesSpaceBefore}\includegraphics[height=2cm]{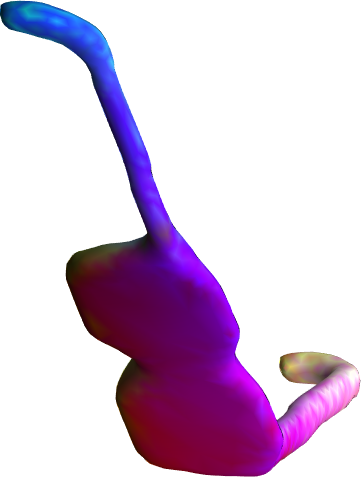}\vspace{\glassesSpaceAfter}\\
    \vspace{\glassesSpaceBefore}\includegraphics[height=2cm]{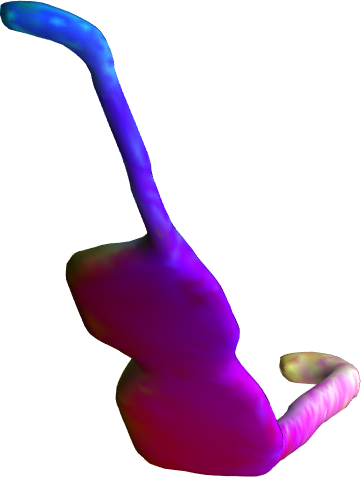}\vspace{\glassesSpaceAfter}\\
  }
  \tabfig{
    \vspace{\teddySpaceBefore}\includegraphics[height=1.8cm]{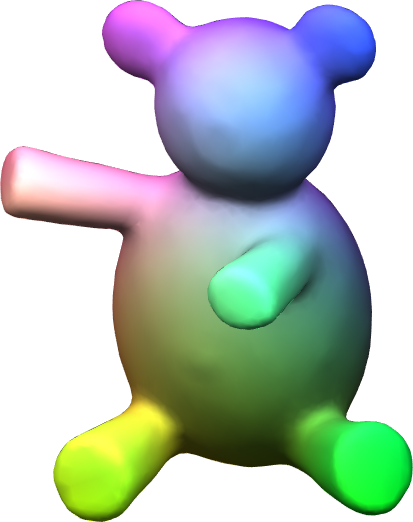}\vspace{\teddySpaceAfter}\\
    \vspace{\teddySpaceBefore}\includegraphics[height=2cm]{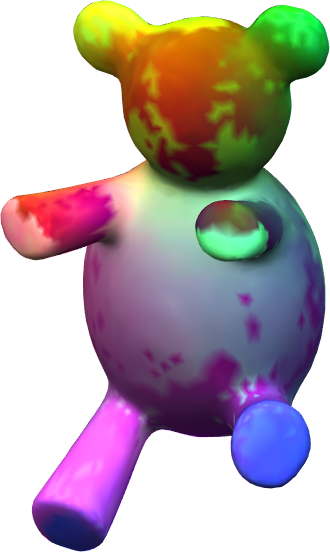}\vspace{\teddySpaceAfter}\\
    \vspace{\teddySpaceBefore}\includegraphics[height=2cm]{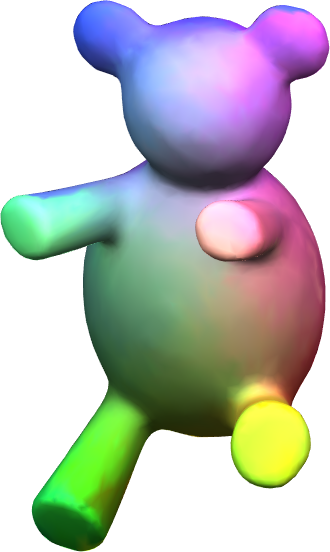}\vspace{\teddySpaceAfter}\\
    \vspace{\teddySpaceBefore}\includegraphics[height=2cm]{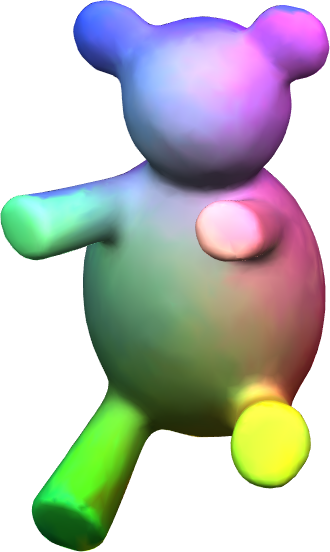}\vspace{\teddySpaceAfter}\\
    \vspace{\teddySpaceBefore}\includegraphics[height=2cm]{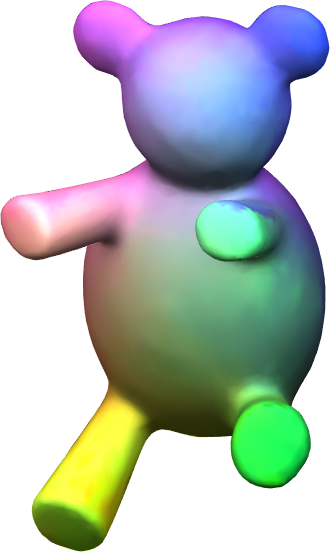}\vspace{\teddySpaceAfter}\\
  }
  \tabfig{
    \vspace{\pigSpaceBefore}\includegraphics[height=1.7cm]{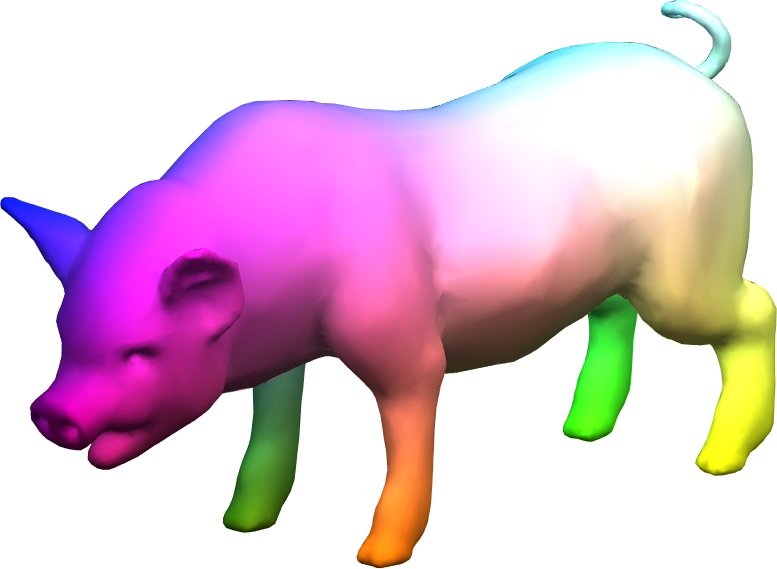}\vspace{\pigSpaceAfter}\\
    \vspace{\pigSpaceBefore}\includegraphics[height=1.7cm]{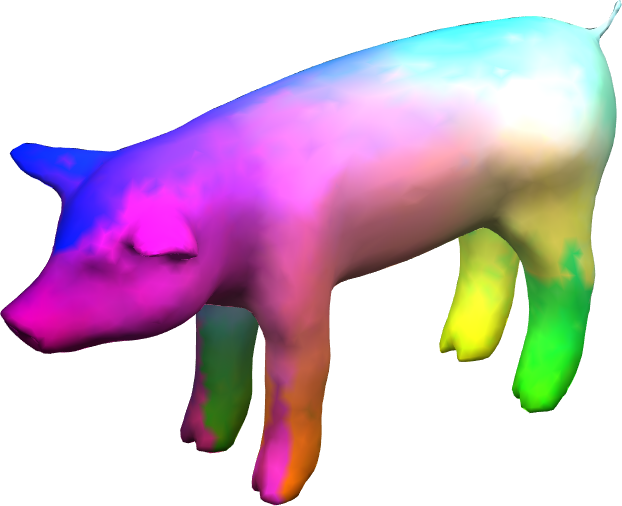}\vspace{\pigSpaceAfter}\\
    \vspace{\pigSpaceBefore}\includegraphics[height=1.7cm]{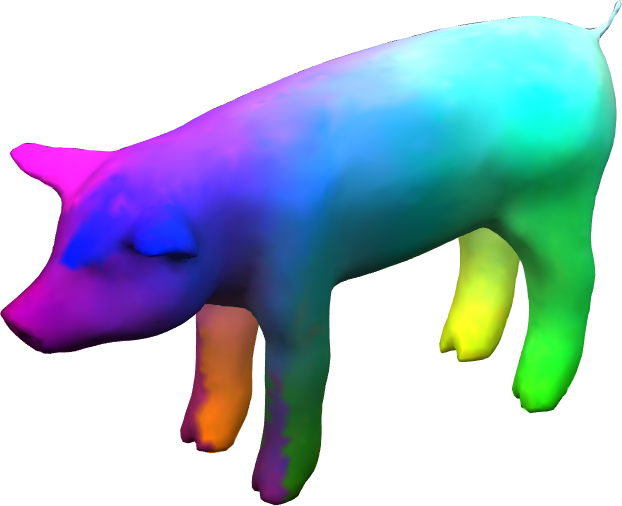}\vspace{\pigSpaceAfter}\\
    \vspace{\pigSpaceBefore}\includegraphics[height=1.7cm]{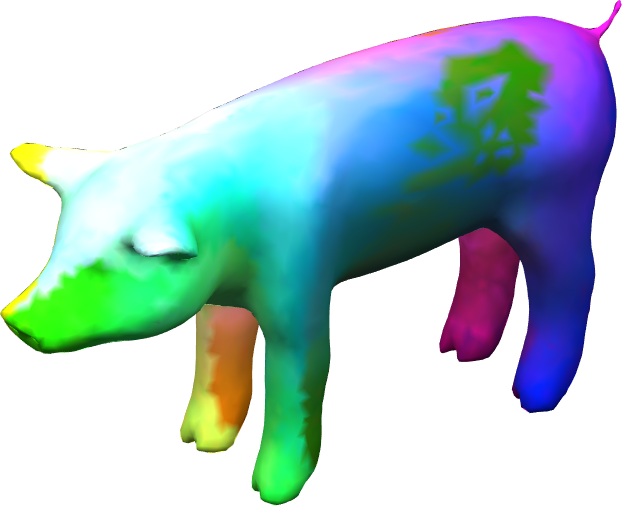}\vspace{\pigSpaceAfter}\\
    \vspace{\pigSpaceBefore}\includegraphics[height=1.7cm]{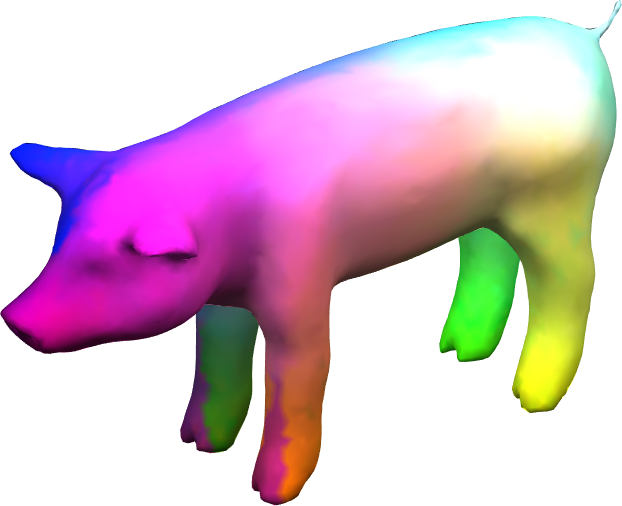}\vspace{\pigSpaceAfter}\\
  }
  \tabfig{
    \vspace{\dogSpaceBefore}\includegraphics[height=1.8cm]{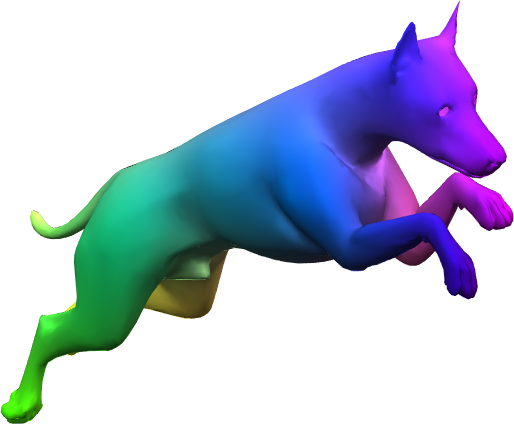}\vspace{\dogSpaceAfter}\\
    \vspace{\dogSpaceBefore}\includegraphics[height=1.8cm]{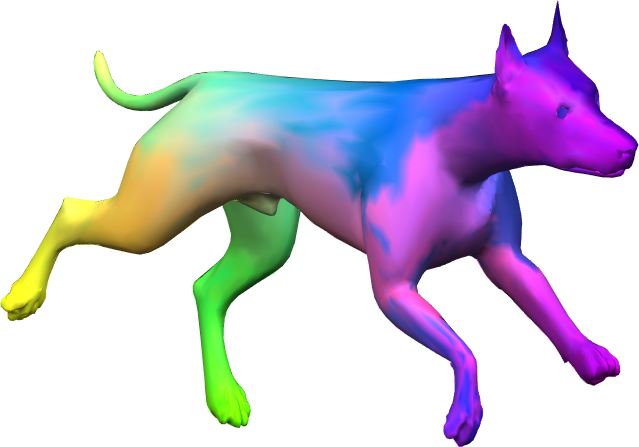}\vspace{\dogSpaceAfter}\\
    \vspace{\dogSpaceBefore}\includegraphics[height=1.8cm]{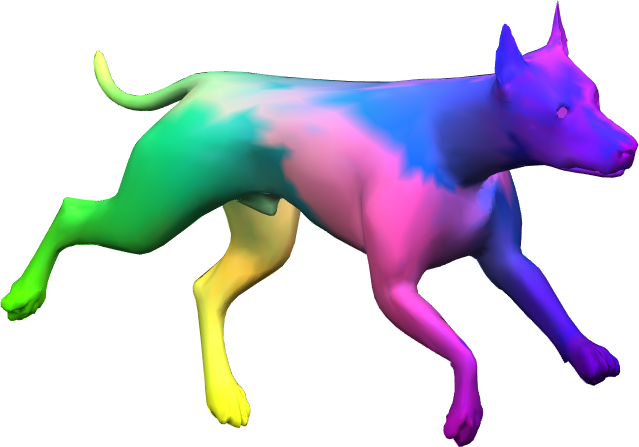}\vspace{\dogSpaceAfter}\\
    \vspace{\dogSpaceBefore}\includegraphics[height=1.8cm]{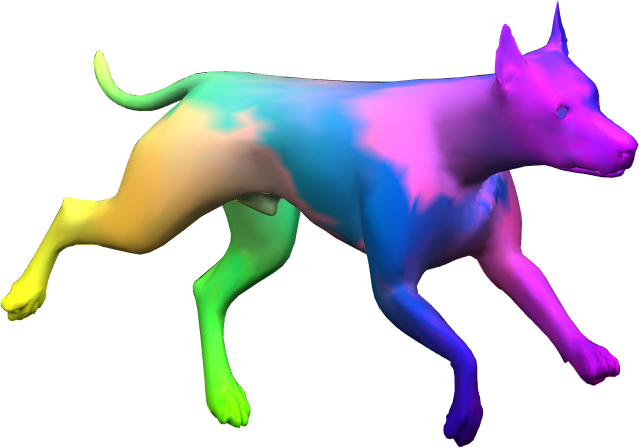}\vspace{\dogSpaceAfter}\\
    \vspace{\dogSpaceBefore}\includegraphics[height=1.8cm]{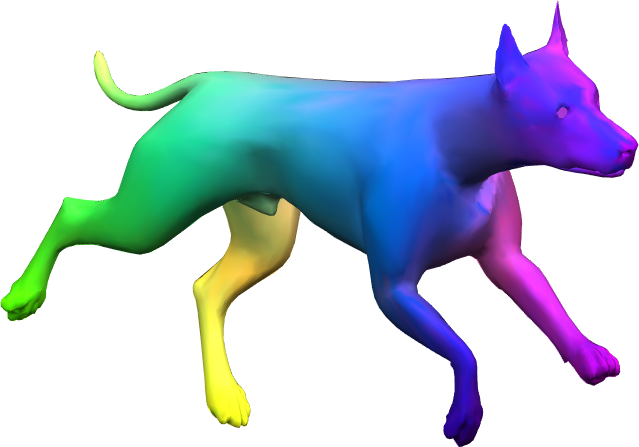}\vspace{\dogSpaceAfter}
  }
  \tabfig{
    \vspace{.2cm}\includegraphics[height=1.8cm]{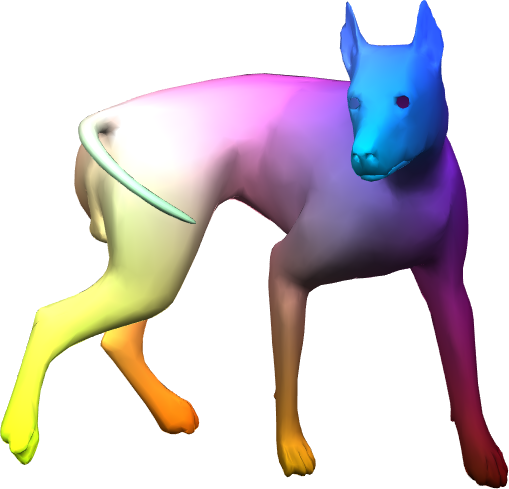}\vspace{.1cm}\\
    \vspace{.2cm}\includegraphics[height=2.2cm]{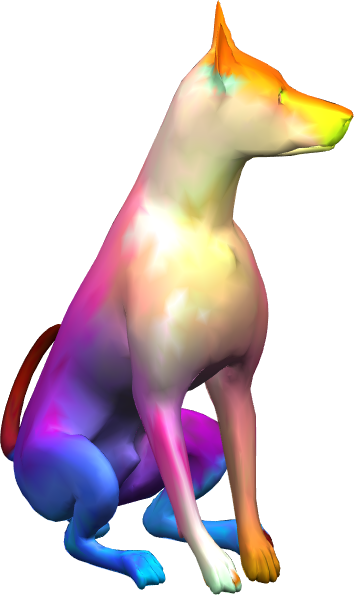}\vspace{.1cm}\\
    \vspace{.2cm}\includegraphics[height=2.2cm]{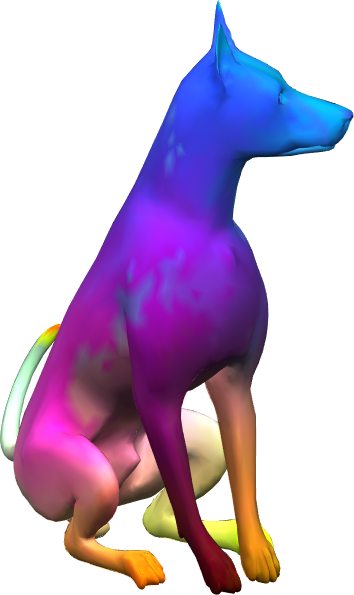}\vspace{.1cm}\\
    \vspace{.2cm}\includegraphics[height=2.2cm]{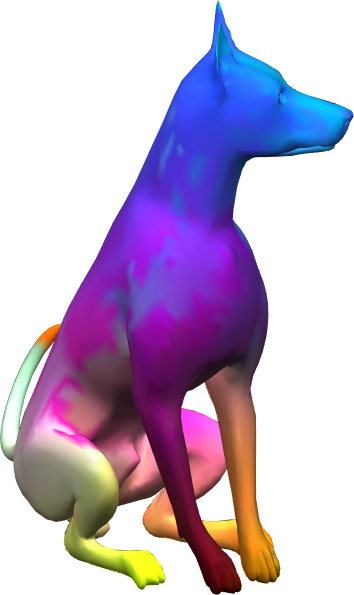}\vspace{.1cm}\\
    \vspace{.2cm}\includegraphics[height=2.2cm]{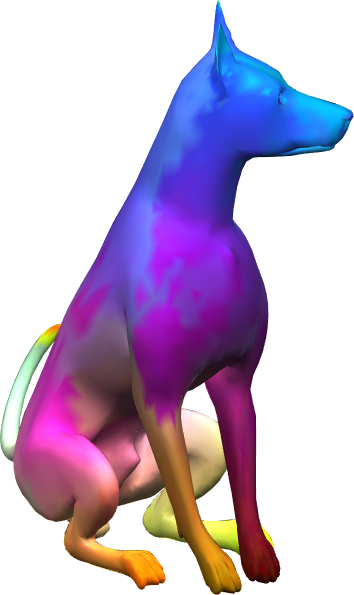}\vspace{.1cm}\\
  }
  \tabfig{
    \includegraphics[height=2.5cm]{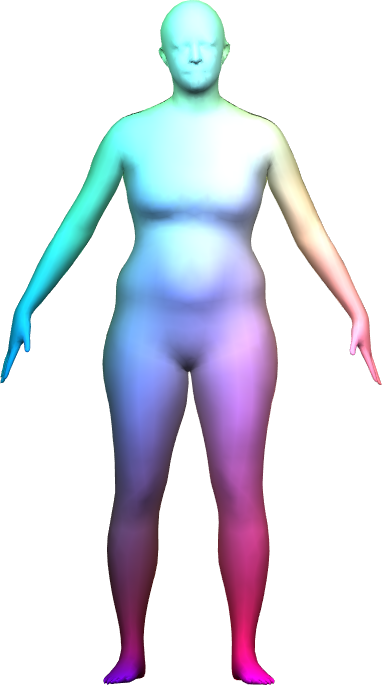}\\
    \includegraphics[height=2.5cm]{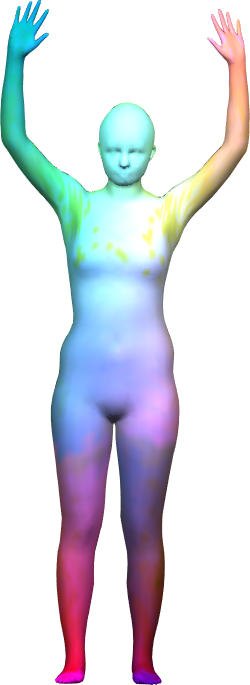}\\
    \includegraphics[height=2.5cm]{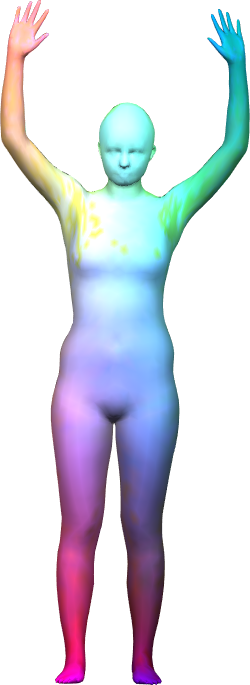}\\
    \includegraphics[height=2.5cm]{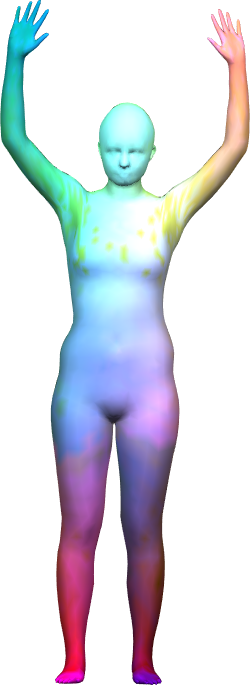}\\
    \includegraphics[height=2.5cm]{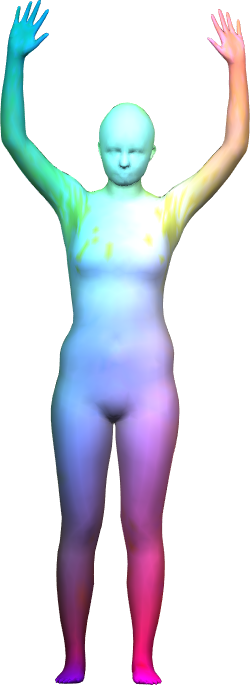}\\
  }
  \tabfig{
    \includegraphics[height=2.5cm]{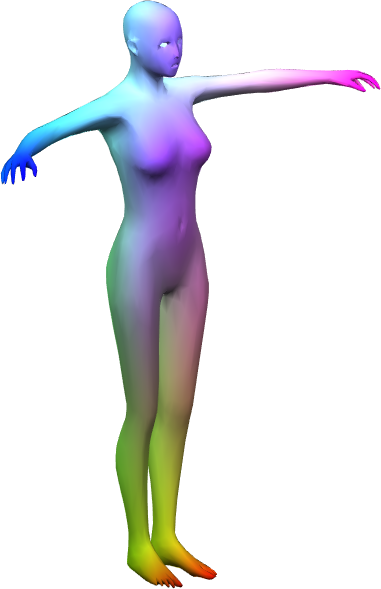}\\
    \includegraphics[height=2.5cm]{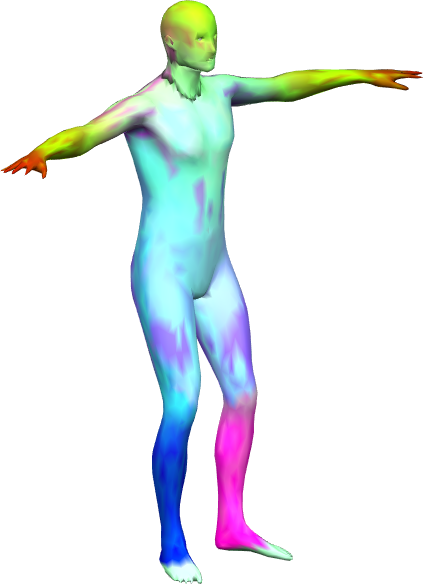}\\
    \includegraphics[height=2.5cm]{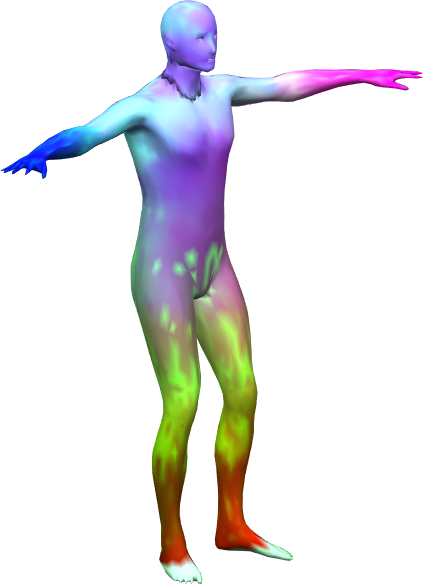}\\
    \includegraphics[height=2.5cm]{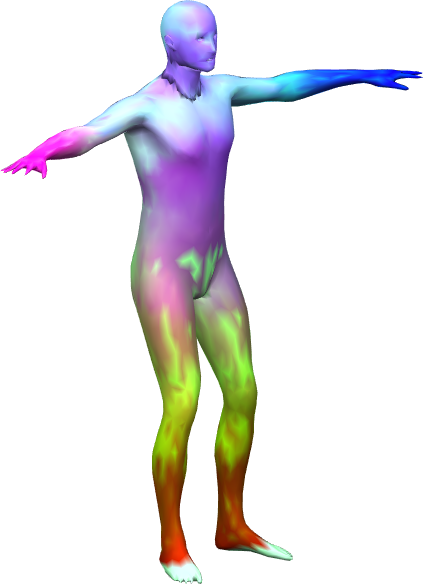}\\
    \includegraphics[height=2.5cm]{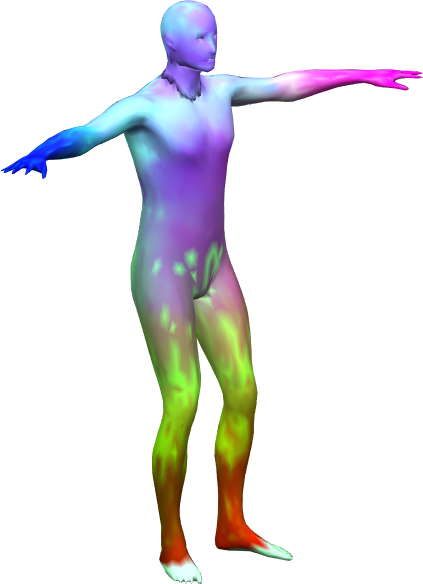}\\
  }
  \tabfig{
    \includegraphics[height=2.5cm]{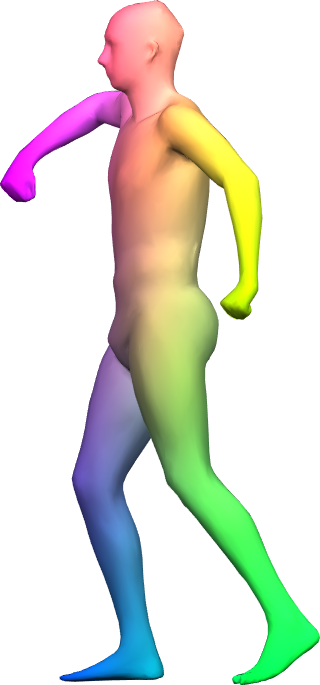}\\
    \includegraphics[height=2.5cm]{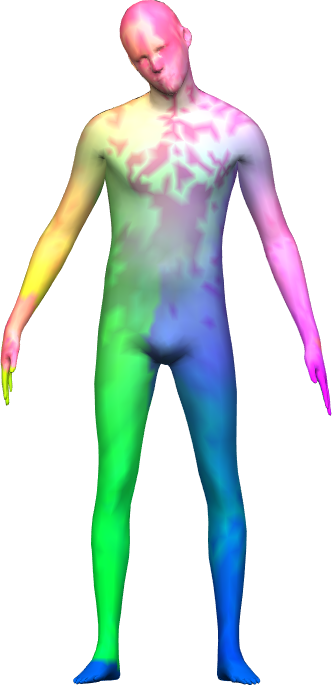}\\
    \includegraphics[height=2.5cm]{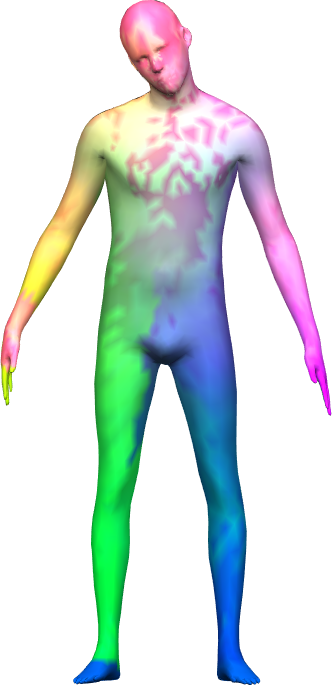}\\
    \includegraphics[height=2.5cm]{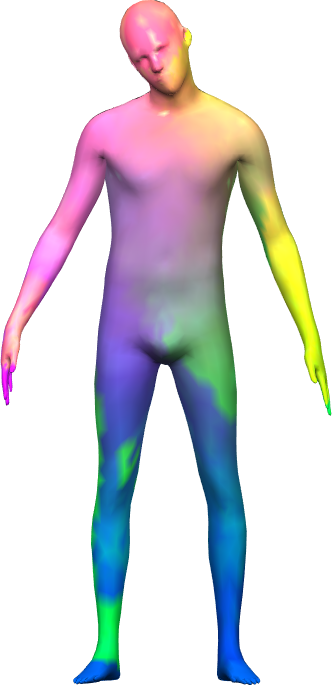}\\
    \includegraphics[height=2.5cm]{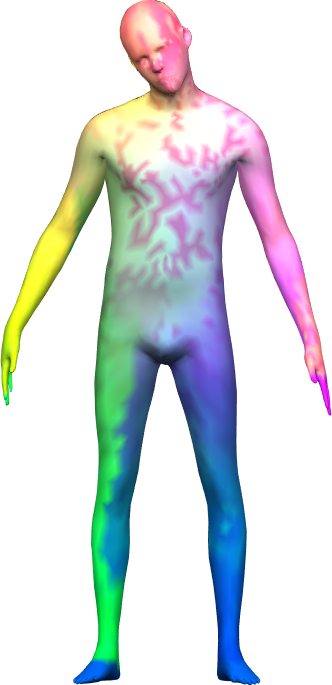}\\
  }
}
}
    \caption{Comparison of different sparse matchings (random, PM-SDP~\cite{maron2016point}, Rodola et al.~\cite{rodola2012game}, MINA) that are used as initialisation for PMF~\cite{vestner2017product} to obtain a dense matching. The first row shows the \emph{reference} shape and the other rows show the colour-coded dense correspondences for the respective methods.}
    \label{fig:denseMatchingResults} 
\end{figure*} 

\subsection{Dense Shape Matching}\label{sec:densematching}
Next, we demonstrate that our method can be used to obtain a suitable initialisation for dense correspondence methods. Since dense non-rigid matching approaches %
are highly initialisation-dependent (even the convex approach~\cite{chen2015robust} requires a good initial alignment, cf.~Sec.~\ref{sec:relatedwork}), it is crucial that they are provided with a good initialisation. 

For these experiments we use
the product manifold filter (PMF)~\cite{vestner2017product} for obtaining a dense matching from a given sparse matching.
To obtain the initial sparse matching, in addition to our MINA approach, we consider a random  matching, a rigid alignment obtained via PM-SDP~\cite{maron2016point}, and the approach by Rodola et al.~\cite{rodola2012game}. Unlike in the previous section, here we extract the sparse points that we want to match based on geodesic farthest point sampling (FPS), which %
obtains an (approximately) uniform sampling of control points on the shapes. %

In Fig.~\ref{fig:denseMatchingResults} we show results for the PMF-based densification for  shapes from the TOSCA dataset~\cite{Bronstein:2008:NGN:1462123} (cat, dog, wolf, human), the SHREC watertight dataset~\cite{watertight} (glasses, teddy, pigs), the FAUST dataset~\cite{Bogo:CVPR:2014} (human) and the SCAPE dataset~\cite{scape} (human). Using a random initialisation (second row) fails in all cases
 and therefore confirms the dependence of PMF to its initialisation. Although  PM-SDP finds a global optimum (of a convex relaxation), the rigid deformation model is too restricted and therefore does not produce reliable dense correspondences for \emph{non-rigid} shape matching (third row). The method by Rodola~et~al.~\cite{rodola2012game} works well for several cases (fourth row), but due to its initialisation-dependence and potential orientation flips it also leads to several wrong matchings. We find that for various types of matching problems, including strong non-rigid deformations (cat in the first column), or inter-object matching (wolf-dog in the second column), our MINA method provides the most reliable initialisation (last row). Although in many cases MINA is able to properly handle self-symmetries, such symmetries form a particular difficulty for all considered methods and therefore may lead to wrong matchings (last two columns). Another difficulty are drastic non-rigid deformations (dog in the fourth last column).

\subsection{Flexibility of MINA Formulation}\label{sec:flex}
Next, we demonstrate the flexibility of our MINA model by addressing several variants of shape matching formulations in a proof-of-concept manner. 
\paragraph{Outlier rejection.}\label{sec:outliers}
So far, we assumed that  there exists a corresponding convex polyhedron on $\shapeY$ for each control point $\shapeX_{\mathcal{I}}$. 
{In order to allow that some control points of $\mathcal{X}_{\mathcal{I}}$ are not matched to a convex polyhedron on $\mathcal{Y}$,}
 we propose to use an outlier rejection mechanism where up to $n_{\text{out}}$ of the control points can remain unmatched. 
 To this end, we 
replace the correspondence term $f_{\text{corr}}$ in~\eqref{eq:corrterm} with 
\begin{align}\label{eq:outlierCorr}
    \tilde{f}_{\text{corr}} := \lambda_\text{c} \, \|  \tau(\shapeX_{\mathcal{I}}) - \matalpha Z + \epsilon  \|\,,
\end{align}
where $\epsilon \in \R^{u{\times}3}$ is a sparse \emph{error variable} with $\|\epsilon\|_0 {\leq} n_{\text{out}}$. Here, we use $\|{\cdot}\|_0$ to denote the row-wise $\ell_0$-norm that counts the total number of non-zero \emph{rows}. To model the $\ell_0$-norm as MIP, we introduce the \emph{outlier indicator} variable $\delta {\in} \{0,1\}^u$, where we impose $\onevec^T_u \delta {\leq} n_{\text{out}}$. Moreover, we make use of the fact that both shapes are spatially bounded, which implies a bounded correspondence error.  
With that, we can enforce sparsity of $\epsilon$ with the linear constraint $-\delta_i M \onevec_3^T {\leq} \epsilon_i {\leq} \delta_i M\onevec_3^T $ for a sufficiently large (positive) number $M$. As such, whenever $\delta_i {=} 1$, the $i$-th control point does not contribute any error towards the term $\tilde{f}_{\text{corr}}$ since $\epsilon_i$ will compensate for the discrepancy between $(\tau(\shapeX_{\mathcal{I}}))_i$ and $(\matalpha Z)_i$. In Fig.~\ref{fig:outlierRejection} we compare our original formulation with the outlier rejection mechanism, which makes it possible to match pairs of shapes even when the control points are inconsistent between both shapes.
\newcommand{\outlierHeight}{1.6cm}
\begin{figure}
\centerline{\scriptsize \hspace{0cm}no outlier rejection\hspace{2cm}with outlier rejection}
\centerline{
\includegraphics[height=\outlierHeight]{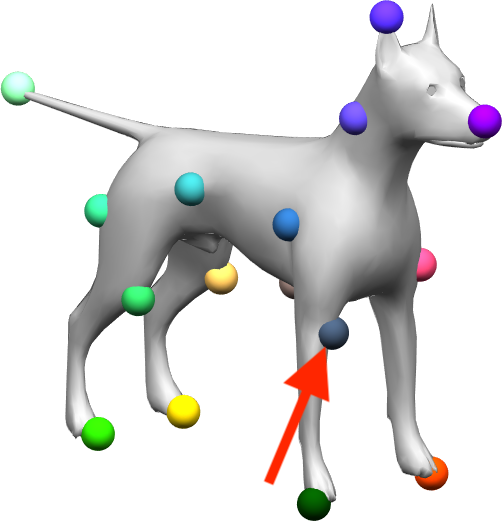}
\includegraphics[height=\outlierHeight]{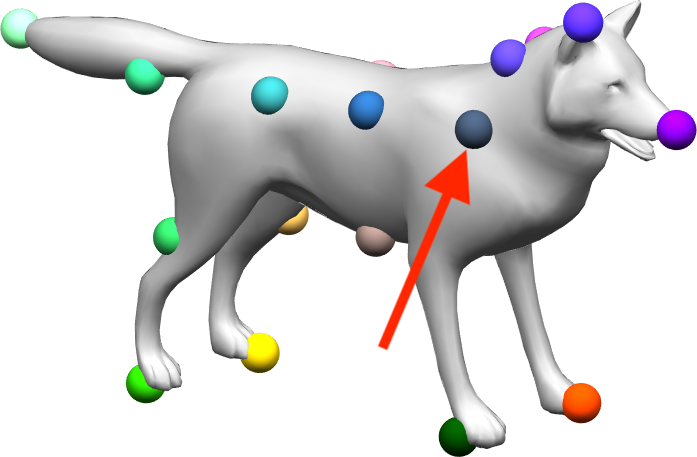}
\rotatebox{90}{\rule{1.8cm}{0.1pt}}
\includegraphics[height=\outlierHeight]{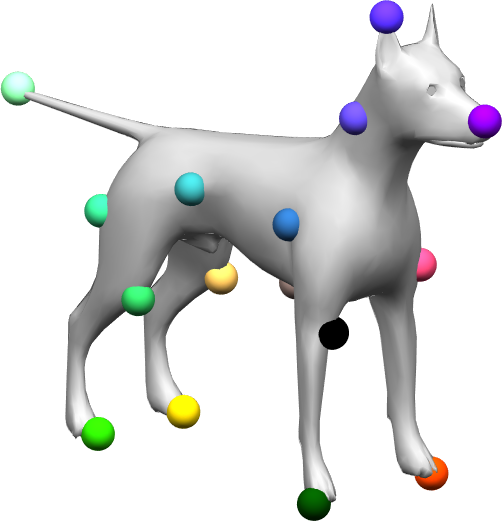}
\includegraphics[height=\outlierHeight]{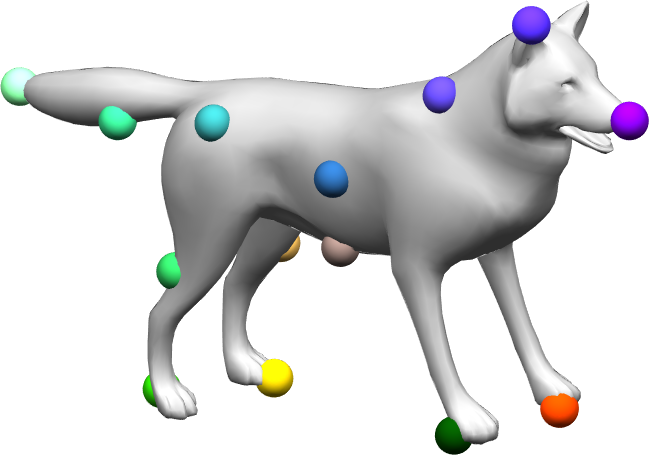}
}
\vspace{-4mm}
{\scriptsize ~~~~~~~~~$\shapeX$ ~~~~~~~~~~~~~~~~~~~~~~~~~~~~~~~~~$\shapeY$~~~~~~~~~~~~~~~~~~~~~~~~~~~~$\shapeX$~~~~~~~~~~~~~~~~~~~~~~~~~~~~~~~~~~$\shapeY$\\}
\vspace{-4mm}
    \caption{Left: without outlier rejection the dog's upper thigh is matched to the wolf's neck (red arrows). Right: our outlier rejection variant~\eqref{eq:outlierCorr} effectively disregards this control point inconsistency (the unmatched point on $\shapeX$ is shown in black).}
    \label{fig:outlierRejection} 
\end{figure}

\paragraph{Shape to point cloud matching.} %
We used MINA for matching a human body mesh (from \cite{scape}) to a real point cloud that we acquired using a \emph{TreedyScan Full Body Scanner}. The raw point cloud was cropped using a manually specified bounding box, downsampled to about 10k points, and denoised. 
The control points where sampled using geodesic FPS, where we used a nearest neighbour graph for computing geodesics (and estimating normals) on the point cloud. For this experiment we enforce that $P$ is an injective matching,~i.e.~we impose $\onevec_u^T P \leq \onevec_v^T$. In Fig.~\ref{fig:teaser} (middle left) we show the resulting matching, which confirms that our method works well in this setting.

\paragraph{Different topologies.} We used MINA for matching two human shapes with different topologies, as shown in Fig.~\ref{fig:teaser} (middle right), where the hands in $\shapeX$ are not touching, whereas the hands in $\shapeY$ are touching, as indicated by the geodesic paths between both hands shown as red lines. Here, we used geodesic FPS to sample the control points. 

\paragraph{Partial shape matching:} We also match a partial shape to a full shape, which we show in Fig.~\ref{fig:teaser} (right). Here, we used geodesic FPS to sample the control points and we enforce that $P$ is an injective matching, as above.

\paragraph{Bounded distortion matching.} {Our formulation also allows to bound}  the maximum distortion of a matching. {This can be implemented by imposing}
linear constraints $P_{st} {+} P_{pq} \leq 1$ for those $s,p{\in}u$ and $t,q{\in}v$ where the geodesic distance between points $s,p$ on $\shapeX$ and points $t,q$ on $\shapeY$ exceed the maximum allowed distortion. With that, at most one of the matchings $P_{st}$ or $P_{pq}$ is allowed.

\section{Discussion \& Limitations}
Although our proposed MINA method has a range of desirable properties, including its high flexibility, its tractability (in practice) due to a low-dimensional matching representation, or its initialisation independence, there are also  open points that we aim to address in the future.
In Sec.~\ref{sec:flex} we demonstrated that MINA enables matching a mesh to real-world point cloud data. 
Considering severely cluttered data,~cf.~\cite{rodola2013scale}, or matching shapes with other data representations (e.g.~polygon soups) are interesting next steps. {A prominent strength of our formulation is that solely using geometric properties already achieves good results. However, additionally} %
incorporating feature descriptors, as commonly done for shape matching, is straightforward and may be useful for further boosting the matching performance.

\paragraph{Scalability.} Our MINA formulation allows to solve non-rigid shape matching problems with $u,v$ being of order $10^2$. Ideally one would be able to address matching problems with a much denser sampling of control points, so that more severe non-rigid deformations can be modelled accurately. 
Although we gained a significant scalability improvement compared to a QAP formulation,~cf.~Fig.~\ref{fig:qaptime}, a further reduction of the computational time would be beneficial. %

\paragraph{Multi matching.} The presented MINA formulation is phrased for matching \emph{pairs} of shapes. We believe that multi matching problems would also benefit from related formulations. One potential way for achieving this is to consider all pairwise matching problems (in a symmetric fashion), and coupling these using cycle-consistency constraints.

\section{Conclusion}
We have presented a convex mixed-integer programming formulation for non-rigid shape matching problems, and we have demonstrated that finding the global optimum is tractable in (most) practical scenarios (see Fig.~\ref{fig:relativeGapStats}). In overall, our formulation comes with a range of benefits: (i) it is more efficient to solve to global optimality compared to the frequently used QAP formulation (Fig.~\ref{fig:qaptime}), (ii) it is initialisation independent, (iii) it is able to obtain suitable initialisations for dense shape matching methods (Sec.~\ref{sec:densematching}), and (iv) it is highly flexible in the type of non-rigid shape matching problems it can handle (Sec.~\ref{sec:flex}). 
Although  MIP formulations are oftentimes evaded for matching problems in computer vision (due to their high computational complexity), in this work we have shown that a suitable problem-specific modelling indeed allows to solve non-rigid shape matching problems as MIP.

$\newline$
\noindent\textbf{Acknowledgement:}
This work was funded by the ERC Consolidator Grant 4DRepLy.

{\small
\bibliographystyle{ieee_fullname}
\bibliography{mina} 

\begin{thebibliography}{10}\itemsep=-1pt

\bibitem{scape}
Dragomir Anguelov, Praveen Srinivasan, Daphne Koller, Sebastian Thrun, Jim
  Rodgers, and James Davis.
\newblock Scape: Shape completion and animation of people.
\newblock In {\em SIGGRAPH}, 2005.

\bibitem{mosek}
MOSEK ApS.
\newblock {\em The MOSEK optimization toolbox for MATLAB manual. Version 9.0.},
  2019.

\bibitem{mosekmodelling}
MOSEK ApS.
\newblock {\em MOSEK Modeling Cookbook. Release 3.2.1}, 2020.

\bibitem{bazaraa1979exact}
Mokhtar~S Bazaraa and Alwalid~N Elshafei.
\newblock An exact branch-and-bound procedure for the quadratic-assignment
  problem.
\newblock {\em Naval Research Logistics Quarterly}, 26(1):109--121, 1979.

\bibitem{bernard2017combinatorial}
Florian Bernard, Frank~R Schmidt, Johan Thunberg, and Daniel Cremers.
\newblock A combinatorial solution to non-rigid 3d shape-to-image matching.
\newblock In {\em CVPR}, 2017.

\bibitem{bernard:2018}
Florian Bernard, Christian Theobalt, and Michael Moeller.
\newblock {DS*: Tighter Lifting-Free Convex Relaxations for Quadratic Matching
  Problems}.
\newblock In {\em CVPR}, 2018.

\bibitem{besl1992method}
Paul~J Besl and Neil~D McKay.
\newblock Method for registration of 3-d shapes.
\newblock In {\em Sensor fusion IV: control paradigms and data structures},
  volume 1611, pages 586--606. International Society for Optics and Photonics,
  1992.

\bibitem{Bogo:CVPR:2014}
Federica Bogo, Javier Romero, Matthew Loper, and Michael~J. Black.
\newblock {FAUST}: Dataset and evaluation for {3D} mesh registration.
\newblock In {\em CVPR}, 2014.

\bibitem{Bronstein:2008:NGN:1462123}
Alexander Bronstein, Michael Bronstein, and Ron Kimmel.
\newblock {\em Numerical Geometry of Non-Rigid Shapes}.
\newblock Springer Publishing Company, Incorporated, 1 edition, 2008.

\bibitem{chen2015robust}
Qifeng Chen and Vladlen Koltun.
\newblock Robust nonrigid registration by convex optimization.
\newblock In {\em CVPR}, 2015.

\bibitem{coughlan2000efficient}
James Coughlan, Alan Yuille, Camper English, and Dan Snow.
\newblock Efficient deformable template detection and localization without user
  initialization.
\newblock {\em Computer Vision and Image Understanding}, 78(3):303--319, 2000.

\bibitem{dai2017global}
Hongkai Dai, Gregory Izatt, and Russ Tedrake.
\newblock Global inverse kinematics via mixed-integer convex optimization.
\newblock {\em The International Journal of Robotics Research}, May 2017.

\bibitem{Dym:2017ue}
Nadav Dym, Haggai Maron, and Yaron Lipman.
\newblock {DS++ - A flexible, scalable and provably tight relaxation for
  matching problems}.
\newblock {\em ACM Transactions on Graphics (TOG)}, 36(6), 2017.

\bibitem{felzenszwalb2005representation}
Pedro~F Felzenszwalb.
\newblock Representation and detection of deformable shapes.
\newblock {\em TPAMI}, 27(2):208--220, 2005.

\bibitem{watertight}
Daniela Giorgi, Silvia Biasotti, and Laura Paraboschi.
\newblock Shape retrieval contest 2007: Watertight models track, 2007.

\bibitem{groueix2019unsupervised}
Thibault Groueix, Matthew Fisher, Vladimir~G Kim, Bryan~C Russell, and Mathieu
  Aubry.
\newblock Unsupervised cycle-consistent deformation for shape matching.
\newblock In {\em Computer Graphics Forum}, volume~38, pages 123--133, 2019.

\bibitem{halimi2019unsupervised}
Oshri Halimi, Or Litany, Emanuele Rodola, Alex~M Bronstein, and Ron Kimmel.
\newblock Unsupervised learning of dense shape correspondence.
\newblock In {\em CVPR}, 2019.

\bibitem{huang2017adjoint}
Ruqi Huang and Maks Ovsjanikov.
\newblock Adjoint map representation for shape analysis and matching.
\newblock In {\em Computer Graphics Forum}, volume~36, pages 151--163, 2017.

\bibitem{Kim11}
Vladimir~G. Kim, Yaron Lipman, and Thomas Funkhouser.
\newblock {Blended Intrinsic Maps}.
\newblock {\em ACM Transactions on Graphics (TOG)}, 30(4), 2011.

\bibitem{kushinsky2019sinkhorn}
Yam Kushinsky, Haggai Maron, Nadav Dym, and Yaron Lipman.
\newblock Sinkhorn algorithm for lifted assignment problems.
\newblock {\em SIAM Journal on Imaging Sciences}, 12(2):716--735, 2019.

\bibitem{lahner2016efficient}
Zorah L\"{a}hner, Emanuele Rodola, Frank~R Schmidt, Michael~M Bronstein, and
  Daniel Cremers.
\newblock Efficient globally optimal 2d-to-3d deformable shape matching.
\newblock In {\em CVPR}, 2016.

\bibitem{le2017alternating}
D~Khu{\^e} L{\^e}-Huu and Nikos Paragios.
\newblock Alternating direction graph matching.
\newblock In {\em CVPR}, 2017.

\bibitem{litany2017deep}
Or Litany, Tal Remez, Emanuele Rodola, Alex Bronstein, and Michael Bronstein.
\newblock Deep functional maps: Structured prediction for dense shape
  correspondence.
\newblock In {\em CVPR}, 2017.

\bibitem{lofberg2004yalmip}
Johan Lofberg.
\newblock Yalmip: A toolbox for modeling and optimization in matlab.
\newblock In {\em ICRA}, 2004.

\bibitem{Loiola:2ua4FrR7}
Eliane~Maria Loiola, Nair Maria~Maia de Abreu, Paulo Oswaldo~Boaventura Netto,
  Peter Hahn, and Tania~Maia Querido.
\newblock {A survey for the quadratic assignment problem.}
\newblock {\em European Journal of Operational Research}, 176(2):657--690,
  2007.

\bibitem{maron2016point}
Haggai Maron, Nadav Dym, Itay Kezurer, Shahar Kovalsky, and Yaron Lipman.
\newblock Point registration via efficient convex relaxation.
\newblock {\em ACM Transactions on Graphics (TOG)}, 35(4):73, 2016.

\bibitem{melzi2019zoomout}
Simone Melzi, Jing Ren, Emanuele Rodol{\`a}, Abhishek Sharma, Peter Wonka, and
  Maks Ovsjanikov.
\newblock Zoomout: Spectral upsampling for efficient shape correspondence.
\newblock {\em ACM Transactions on Graphics (TOG)}, 38(6), 2019.

\bibitem{mena2018learning}
Gonzalo Mena, David Belanger, Scott Linderman, and Jasper Snoek.
\newblock Learning latent permutations with gumbel-sinkhorn networks.
\newblock In {\em ICLR}, 2018.

\bibitem{Munkres:1957ju}
James Munkres.
\newblock {Algorithms for the Assignment and Transportation Problems}.
\newblock {\em Journal of the Society for Industrial and Applied Mathematics},
  5(1):32--38, Mar. 1957.

\bibitem{myronenko2010point}
Andriy Myronenko and Xubo Song.
\newblock Point set registration: Coherent point drift.
\newblock {\em TPAMI}, 32(12):2262--2275, 2010.

\bibitem{olsson2008branch}
Carl Olsson, Fredrik Kahl, and Magnus Oskarsson.
\newblock Branch-and-bound methods for euclidean registration problems.
\newblock {\em TPAMI}, 31(5):783--794, 2008.

\bibitem{ovsjanikov2012functional}
Maks Ovsjanikov, Mirela Ben-Chen, Justin Solomon, Adrian Butscher, and Leonidas
  Guibas.
\newblock Functional maps: a flexible representation of maps between shapes.
\newblock {\em ACM Transactions on Graphics (TOG)}, 31(4):30, 2012.

\bibitem{qi2017pointnet}
Charles~R Qi, Hao Su, Kaichun Mo, and Leonidas~J Guibas.
\newblock Pointnet: Deep learning on point sets for 3d classification and
  segmentation.
\newblock In {\em CVPR}, 2017.

\bibitem{ren2018continuous}
Jing Ren, Adrien Poulenard, Peter Wonka, and Maks Ovsjanikov.
\newblock Continuous and orientation-preserving correspondences via functional
  maps.
\newblock In {\em SIGGRAPH Asia}, 2018.

\bibitem{rendl1994quadratic}
F Rendl, P Pardalos, and H Wolkowicz.
\newblock The quadratic assignment problem: A survey and recent developments.
\newblock In {\em Proceedings of the DIMACS workshop on quadratic assignment
  problems}, volume~16, pages 1--42, 1994.

\bibitem{rodola2013scale}
Emanuele Rodol{\`a}, Andrea Albarelli, Filippo Bergamasco, and Andrea Torsello.
\newblock A scale independent selection process for 3d object recognition in
  cluttered scenes.
\newblock {\em IJCV}, 102(1-3):129--145, 2013.

\bibitem{rodola2012game}
Emanuele Rodola, Alex~M Bronstein, Andrea Albarelli, Filippo Bergamasco, and
  Andrea Torsello.
\newblock A game-theoretic approach to deformable shape matching.
\newblock In {\em CVPR}, 2012.

\bibitem{rodola2014dense}
Emanuele Rodol{\`a}, Samuel Rota~Bulo, Thomas Windheuser, Matthias Vestner, and
  Daniel Cremers.
\newblock Dense non-rigid shape correspondence using random forests.
\newblock In {\em CVPR}, 2014.

\bibitem{roufosse2018unsupervised}
Jean-Michel Roufosse, Abhishek Sharma, and Maks Ovsjanikov.
\newblock Unsupervised deep learning for structured shape matching.
\newblock In {\em ICCV}, 2019.

\bibitem{Schellewald:2005up}
Christian Schellewald and Christoph Schn{\"o}rr.
\newblock {Probabilistic subgraph matching based on convex relaxation}.
\newblock In {\em EMMCVPR}, 2005.

\bibitem{schoenemann2009combinatorial}
Thomas Schoenemann and Daniel Cremers.
\newblock A combinatorial solution for model-based image segmentation and
  real-time tracking.
\newblock {\em TPAMI}, 32(7):1153--1164, 2009.

\bibitem{schonemann1966generalized}
Peter~H Sch{\"o}nemann.
\newblock A generalized solution of the orthogonal procrustes problem.
\newblock {\em Psychometrika}, 31(1):1--10, 1966.

\bibitem{sorkine2007rigid}
Olga Sorkine and Marc Alexa.
\newblock As-rigid-as-possible surface modeling.
\newblock In {\em Symposium on Geometry processing}, 2007.

\bibitem{sumner2004deformation}
Robert~W Sumner and Jovan Popovi{\'c}.
\newblock Deformation transfer for triangle meshes.
\newblock {\em ACM Transactions on graphics (TOG)}, 23(3):399--405, 2004.

\bibitem{sumner2007embedded}
Robert~W Sumner, Johannes Schmid, and Mark Pauly.
\newblock Embedded deformation for shape manipulation.
\newblock {\em ACM Transactions on Graphics (TOG)}, 26(3):80, 2007.

\bibitem{swoboda2017b}
Paul Swoboda, Carsten Rother, Hassan~Abu Alhaija, Dagmar Kainmüller, and
  Bogdan Savchynskyy.
\newblock Study of lagrangean decomposition and dual ascent solvers for graph
  matching.
\newblock In {\em CVPR}, 2017.

\bibitem{van2011survey}
Oliver Van~Kaick, Hao Zhang, Ghassan Hamarneh, and Daniel Cohen-Or.
\newblock A survey on shape correspondence.
\newblock In {\em Computer Graphics Forum}, volume~30, pages 1681--1707. Wiley
  Online Library, 2011.

\bibitem{vestner2017efficient}
Matthias Vestner, Zorah L{\"a}hner, Amit Boyarski, Or Litany, Ron Slossberg,
  Tal Remez, Emanuele Rodola, Alex Bronstein, Michael Bronstein, Ron Kimmel,
  and Daniel Cremers.
\newblock Efficient deformable shape correspondence via kernel matching.
\newblock In {\em 3DV}, 2017.

\bibitem{vestner2017product}
Matthias Vestner, Roee Litman, Emanuele Rodol{\`a}, Alex Bronstein, and Daniel
  Cremers.
\newblock Product manifold filter: Non-rigid shape correspondence via kernel
  density estimation in the product space.
\newblock In {\em CVPR}, 2017.

\bibitem{vielma2011modeling}
Juan~Pablo Vielma and George~L Nemhauser.
\newblock Modeling disjunctive constraints with a logarithmic number of binary
  variables and constraints.
\newblock {\em Mathematical Programming}, 128(1-2):49--72, 2011.

\bibitem{windheuser2011geometrically}
Thomas Windheuser, Ulrich Schlickewei, Frank~R Schmidt, and Daniel Cremers.
\newblock Geometrically consistent elastic matching of 3d shapes: A linear
  programming solution.
\newblock In {\em ICCV}, 2011.

\bibitem{yang2013go}
Jiaolong Yang, Hongdong Li, and Yunde Jia.
\newblock Go-icp: Solving 3d registration efficiently and globally optimally.
\newblock In {\em CVPR}, 2013.

\bibitem{Zhou:2016ty}
Feng Zhou and Fernando De~la Torre.
\newblock {Factorized Graph Matching}.
\newblock {\em TPAMI}, 38(9):1774--1789, 2016.

\end{thebibliography}
}

\clearpage
\part*{Supplementary Material}
\appendix
\section{Obtaining Convex Polyhedra on $\shapeY$}\label{sec:cvxpoly}

Given the $j$-th control point of $\shapeY_{\mathcal{J}}$, we obtain its associated convex polyhedron using a neighhourhood propagation strategy.
To this end, we define a \emph{planarity criterion} using the maximum of the \emph{mean absolute deviation} (MAD) of the surface normals at the points in $Z_j$. For a given matrix $N \in \R^{n {\times} 3}$  and its  column mean $\overline{N} \in \R^{1{\times}3}$, the MAD is defined as $\operatorname{mad}(N) := \frac{1}{n}\sum_i |N_i - \overline{N}|$. As such, starting with $t{=}0$, we consider the vertices of the $t$-ring of the $j$-th vertex as $Z_j$, where we increase $t$ as long as $\max(\operatorname{mad}(N_j^t)) \leq \eta$. Here, $N_j^t$ denotes the  matrix of the normals of the $t$-ring of the $j$-th vertex and $\eta$ specifies a user-defined threshold. Once we have determined the largest $t$ such that the $t$-ring neighborhood is sufficiently planar (below the threshold $\eta$), we discard all points in the rows of $Z_j$ that are interior vertices of the convex polygon defined by $Z_j$ (as they are redundant). In Fig.~\ref{fig:ptToCvxPoly} (right) we show so-obtained convex polyhedra. 

\section{Piece-wise Linear Approximation of $\SO(3)$ Constraints}
The constraint  $R {\in} \SO(3)$ can be expressed as the orthogonality constraint $R^TR {=} \matI_3$ in combination with $R_1 {\times} R_2 {=} R_3$. 
Hence, the constraint $R {\in} \SO(3)$ comprises  exactly $6$ quadratic equality constraints, which form a non-convex set. In order to define a piece-wise linear approximation we use \emph{specially-ordered set of type 2} (sos2) {variables}. An sos2 variable is a non-negative vector where at most two consecutive element can be non-zero. {With that, such a variable allows to encode } a non-convex quadratic function in terms of a piece-wise linear one, so that in the end
 all quadratic constraints become linear, and the sos2 constraints are imposed based on (few) binary variables.

For illustrative purposes, we will now provide a simple example for a piece-wise linear approximation of a quadratic function. Let us consider the function $h(x) = x^2$ on the interval $[-1,1]$. First, we split the domain into $b$ bins, so that we evaluate $x^2$ at these $b$ discrete positions, and then compute all values that fall in-between the sampled points as linear approximation between its two neighbour sample points. 
 Let $b=4$, and let $\phi = [-1, -0.5, 0, 0.5, 1]^T$ be a vector that contains the discretised domain, so that $\phi^2 = [1, 0.25, 0, 0.25, 1]^T$ defines the elementwise square of $\phi$. Moreover, let $\lambda \in \R^{b{+}1}$ be a non-negative sos2 variable that sums to one (as mentioned, sos2 means that only two consecutive elements can be non-zero). Then, we can approximate 
 \begin{align}\label{eq:eqsos2}
 	h(x) \approx \lambda^T \phi^2 \quad \text{ for } \quad x = \lambda^T \phi\,.
 \end{align}
 For example, for $x = 0.75$, we obtain the sos2 variable $\lambda = [0,0,0,0.5,0.5]^T$ (since $x = 0.75 = \lambda^T \phi$). With that, we obtain $h(0.75) =  0.5625 \approx 0.625 = \lambda^T \phi^2$. The important property is that~\eqref{eq:eqsos2} allows to approximate the quadratic function $h(\cdot)$ based on a representation that is \emph{linear} in the variables $x$ and $\lambda$.
In addition to \cite{dai2017global} and \cite{vielma2011modeling}, we refer the interested reader to \cite[Ch. 9.1.11]{mosekmodelling}\footnote{also available online at\\\url{https://docs.mosek.com/modeling-cookbook/mio.html\#continuous-piecewise-linear-functions}}, where sos2 constraints as well as the idea of using a logarithmic Gray encoding are explained.

\section{Search Space Reduction} \label{sec:ssred}
\begin{figure}
\centerline{ 
  \begin{tabular}{@{}c@{}}{\includegraphics[width=0.29\linewidth]{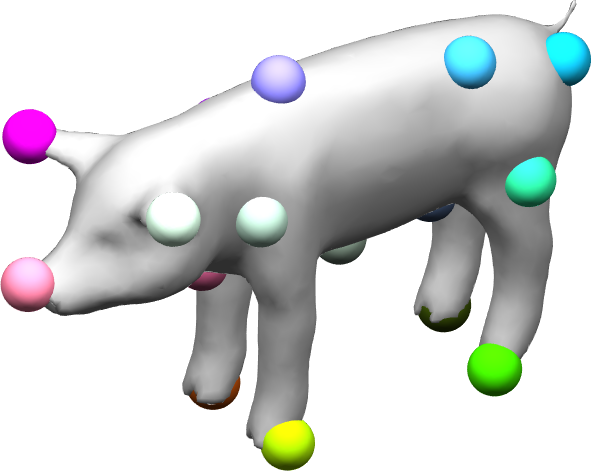}}\\$\shapeX$\end{tabular}
~~~
  \begin{tabular}{@{}c@{}}{\includegraphics[width=0.3\linewidth]{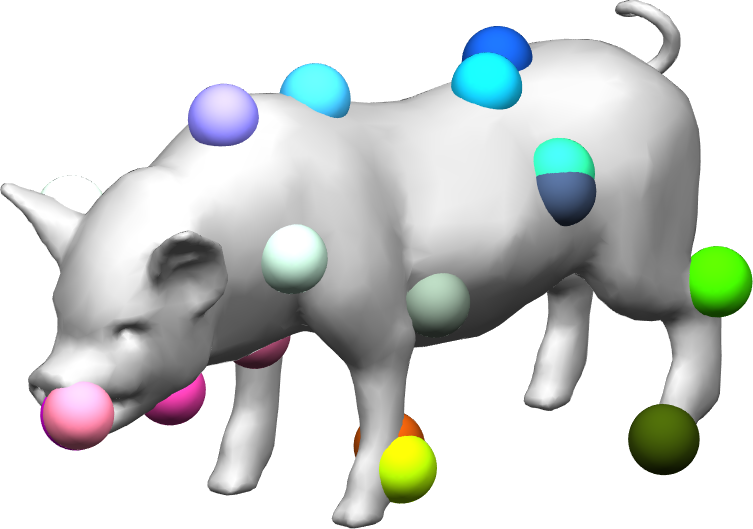}}\\$\shapeY$ (ADG)\end{tabular}
  ~~~
  \begin{tabular}{@{}c@{}}{\includegraphics[width=0.3\linewidth]{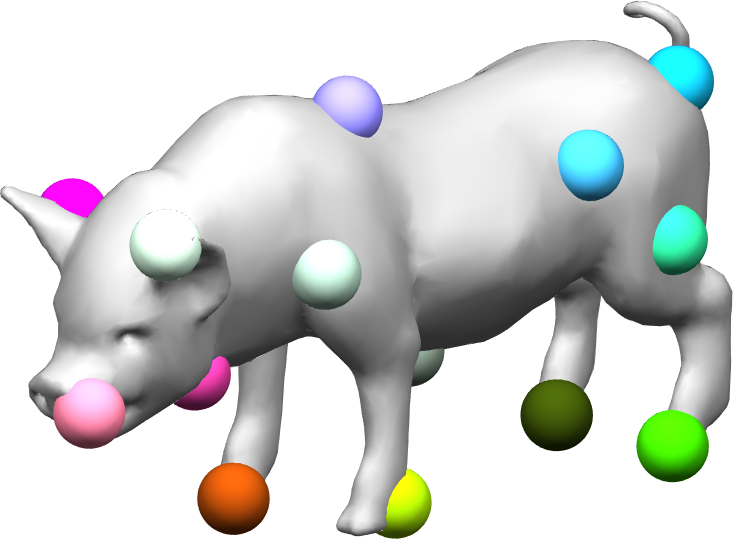}}\\$\shapeY$  (ours)\end{tabular}
}
    \caption{Shape $\shapeX$ (left) is matched to $\shapeY$, where a search space reduction using ADG~\cite{maron2016point} leads to wrong matchings (middle), whereas ours produces correct correspondences (right).}
    \label{fig:percentile} 
\end{figure}

In addition to using a logarithmic encoding of the $\SO(3)$ discretisation variables, we also impose further constraints upon the matching matrix $P$, so that  the size of the overall search space can be reduced. A similar idea has also been pursued in~\cite{maron2016point}, where a scalar criterion based on the average geodesic distance (ADG) was used. In contrast, rather than using a single scalar value for each vertex, we propose to leverage a more powerful approach that considers more descriptive statistics of geodesic distances, see Fig.~\ref{fig:percentile}.
To this end, for each control point we compute $n_{\text{prctile}}$ evenly spaced percentiles from $0$ to $100\%$ of the geodesic distance from this control point to all other points. Let $\gamma^{\shapeX} \in \R^{u{\times}n_{\text{prctile}}}$ and $\gamma^{\shapeY} \in \R^{v{\times}n_{\text{prctile}}}$ denote the so-obtained percentile matrices, where the columns are the ordered percentiles  from $0$ to $100\%$. 
As such, the matrices $\gamma^{\shapeX}$ and $\gamma^{\shapeY}$ can be seen as features of the respective shapes extracted at the control points.  Whenever two control points $i \in [u], j \in [v]$ correspond to each other, the features $\gamma^{\shapeX}_i$ and $\gamma^{\shapeY}_j$ should be similar, so that $d_{ij} {:=} \|\gamma^{\shapeX}_i-\gamma^{\shapeY}_j\|$ is small. 
Based on this observation, we use the feature distances $[d_{ij}]_{i,j}$ and sequentially solve $n_{\text{LAP}}$ linear assignment problems (LAP)~\cite{Munkres:1957ju} to match features. The idea of solving a sequence of LAPs is to not only find the single best matching $P_1$ of features, but rather finding multiple solutions $P_1,\ldots,P_{n_{\text{LAP}}}$, so that the nonzero elements in $P_{\text{all}} = \sum_{\ell=0}^{n_{\text{LAP}}} P_{\ell}$ define the allowed matchings in $P$. Here, the matrix $P_\ell$ is obtained by performing a feature matching using $[d_{ij}]_{i,j}$ when forbidding all previous matchings  $P_1,\ldots,P_{\ell{-}1}$. As such, when optimising MINA, we constrain all elements of $P$ to be zero for those elements where $P_{\text{all}}$ is zero. Using this procedure is advantageous over simple thresholding of $[d_{ij}]_{i,j}$, since on the one hand feasibility is guaranteed, and on the other hand the number of allowed matchings is equal for all control points.  

\section{Further Implementation Details}
We have implemented MINA in the  optimisation modelling toolbox Yalmip~\cite{lofberg2004yalmip}, which uses the conic mixed-integer branch and bound solver MOSEK~\cite{mosek} as backend (with default parameters). In all experiments we used $\lambda_{\text{c}}{=}4, \lambda_{\text{r}}{=}1$ and $\lambda_{\text{s}}{=}0.5$, 
where we account for different problem sizes by   multiplying each $\lambda_{\bullet}$ with $\frac{1}{\sqrt{\#}}$, where $\#$ denotes the total number of elements that the norm is applied to.
We set the weights $\omega_e$ for the smoothness term to $\omega_e {=} \frac{d_e}{\sum_{e \in \mathcal{E}} d_e}$, where for $e=(p,q)$ by $d_e$ we denote the length of the common edge of triangles $p,q$. With that, we achieve that the deformation  of two adjacent triangles $p,q$ is more flexible when their common edge is small. We set the planarity threshold to $\eta{=}\frac{1}{2}$. For keeping the number of variables small, for each convex polyhedron $Z_j$ we only keep the respective control point as well as four additional points obtained via farthest point sampling (FPS) using geodesic distances as metric. Note that this results in convex polyhedra that are either a single point (if none of the $t$-rings of the $j$-th control point satisfies the planarity criterion), or $Z_j$ is a $5{\times}3$ matrix. %
Since the non-rigid deformation induced by a sparse set of matched control points is relatively coarse, rather than modelling $\tau$ with the original mesh resolution we use downsampled meshes with about $300$ faces, similarly as in~\cite{sumner2007embedded}. We set $n_{\text{LAP}} {=} 5$,  $n_{\text{prctile}} {=} \min(n_{\shapeX}, n_{\shapeY})$, $M{=}0.2$ and use $b{=}4$ bins for the $\SO(3)$ discretisation.

Next, we provide additional details on shape to point cloud matching and the relation between partial shape matching and outlier rejection.

\paragraph{Shape to point cloud matching.}
The main difference when $\shapeY$ is represented as a point cloud rather than a mesh is that we need to use a different approach for computing geodesic distances and normals (required for sampling control points, for the definition of the convex polyhedra as described in Sec.~\ref{sec:cvxpoly}, and for the search space reduction described in Sec.~\ref{sec:ssred}). In our case we compute geodesic distances and normals based on a nearest neighbour graph, where we use the $3$ nearest neighbours. After this information is obtained, the overall optimisation problem is equivalent to the one when $\shapeY$ is a mesh, since the only information of $\shapeY$ that is explicitly used in our optimisation problem formulation are the convex polyhedra.

\paragraph{Relation between partial shape matching and outlier rejection.} In our considered \emph{partial shape matching} setting we match \emph{all} control points of the partial shape $\mathcal{X}$ to the full shape $\mathcal{Y}$. This is in contrast to our \emph{outlier rejection} setting, where we allow that some control points of $\mathcal{X}$ are not matched to $\mathcal{Y}$. However, although for partial shape matching we do not use outlier rejection, we mention that principally it could be used for matching a full shape to a partial one.

\clearpage
\section{Additional TOSCA Results}
\begin{figure}[h]
       \centerline{ \includegraphics[width=1\linewidth]{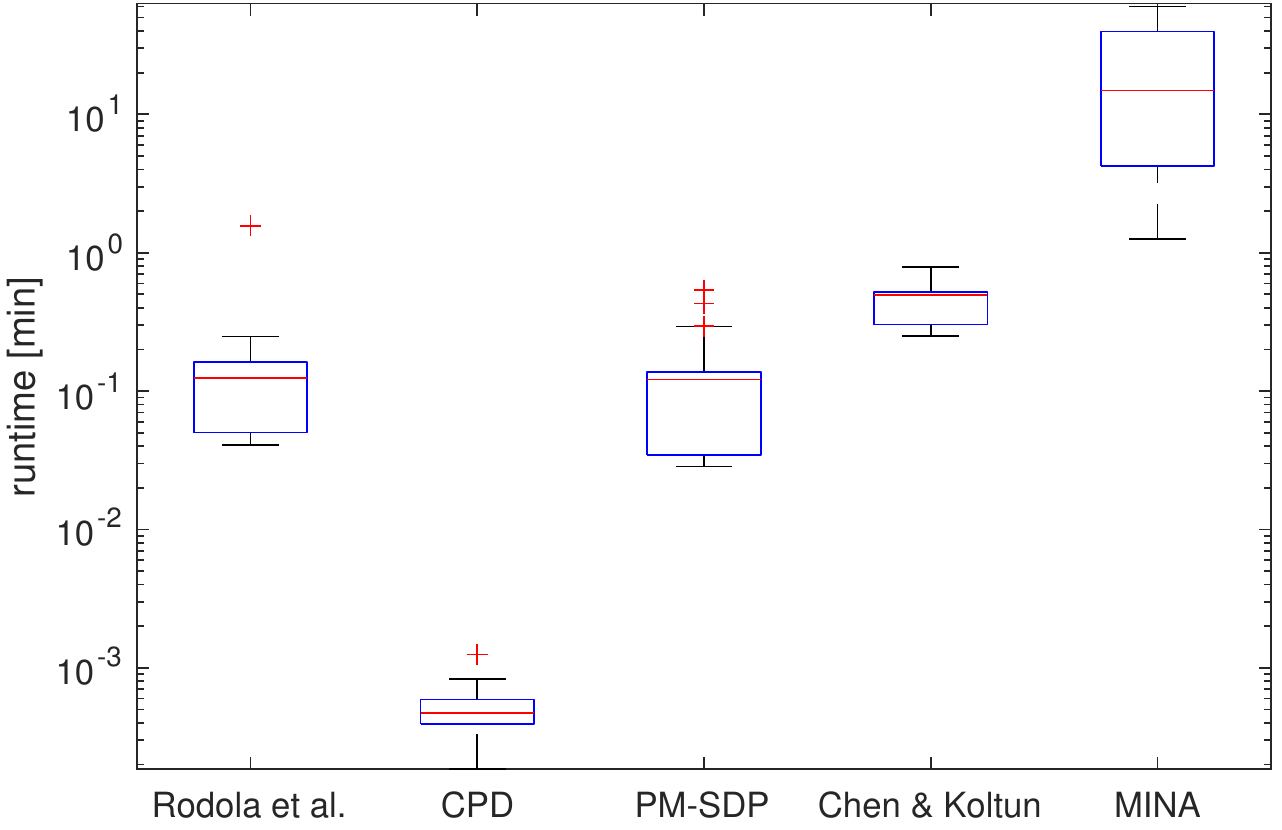}
       }
       \vspace{-1mm}
    \caption{Runtime statistics for the TOSCA dataset. Note that the vertical axis is shown in log-scale.}
    \label{fig:toscaRuntime} 
\end{figure}

In Fig.~\ref{fig:toscaRuntime} we report runtime statistics over all $71$ shape matching instances from the TOSCA datasets for all considered methods. On this dataset, the median processing time of our method is ${\approx}15$min, whereas the other methods require less than one minute.

In Fig.~\ref{fig:qualitativeResultsToscaSupp} we present further results where also the deformed shape $\tau(\shapeX)$ is shown.
\newcommand{\heightA}{4.7cm}
\newcommand{\heightB}{4.5cm}

\begin{figure*}[t!]
\centerline{
\scalebox{0.67}{
	\tabfig{
    \includegraphics[height=\heightA]{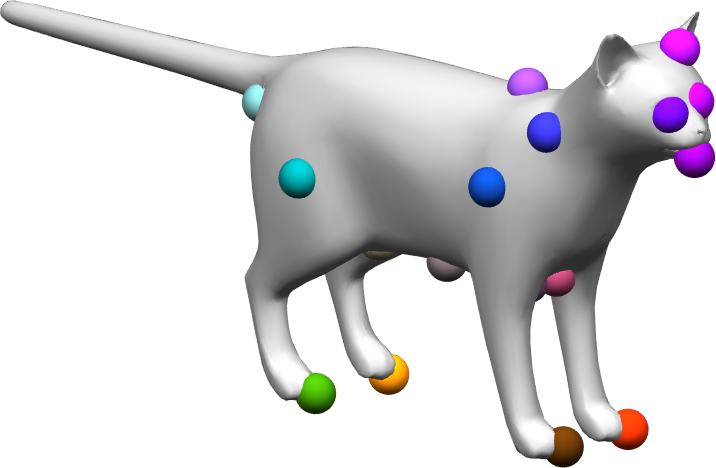}\\
    \includegraphics[height=\heightB]{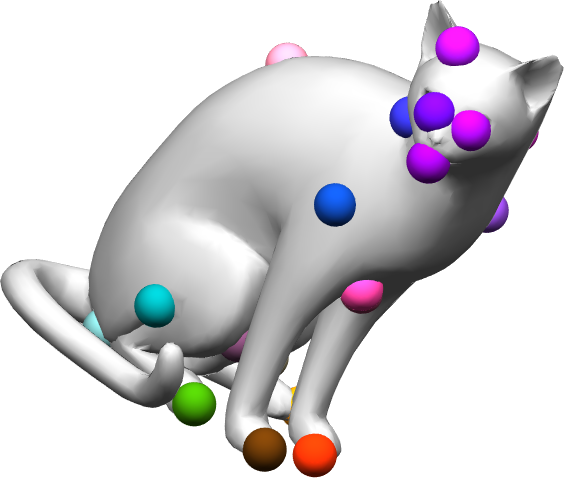}\\
    \includegraphics[height=\heightA]{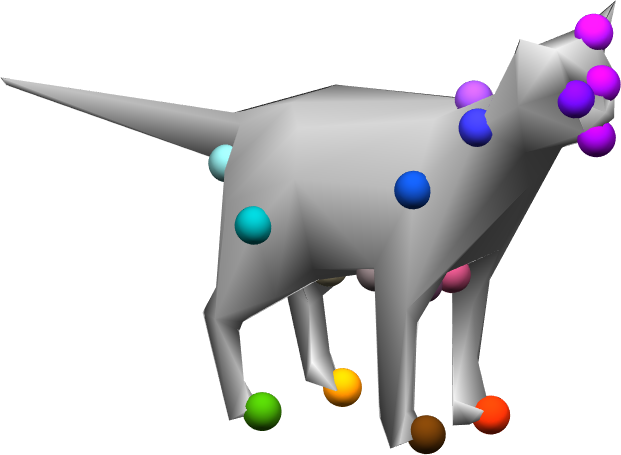}\\
    }
    \tabfig{
    \includegraphics[height=\heightA]{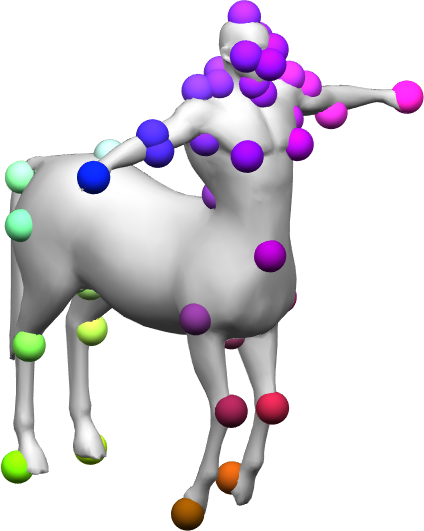}\\
    \includegraphics[height=\heightB]{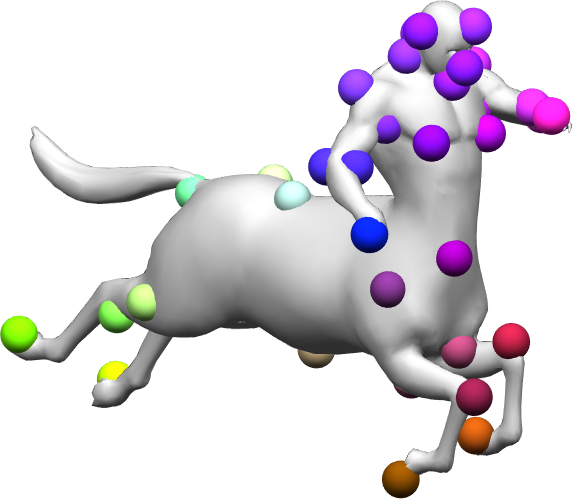}\\
    \includegraphics[height=\heightA]{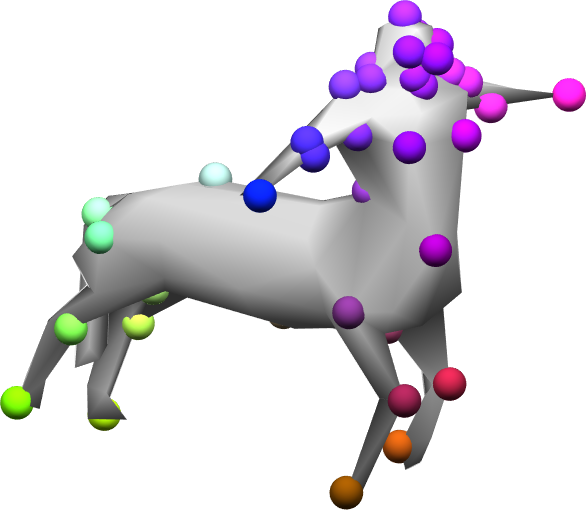}\\
    }
    \tabfig{
    \includegraphics[height=\heightA]{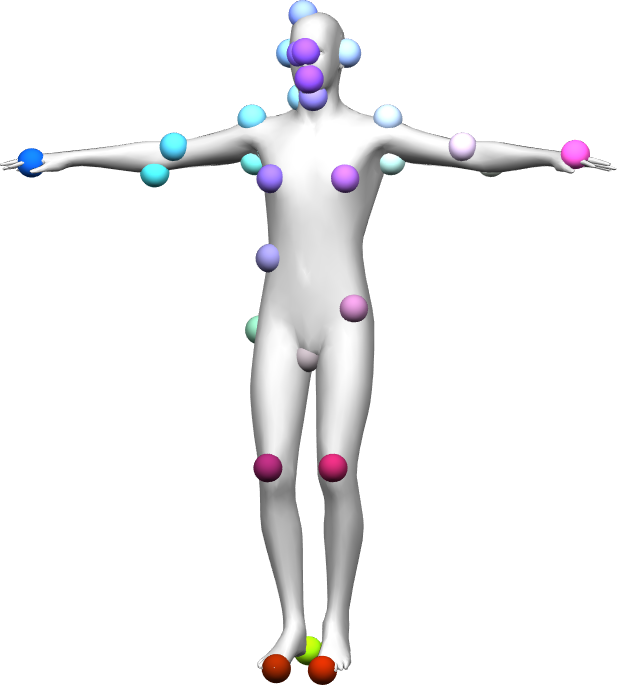}\\
    \includegraphics[height=\heightB]{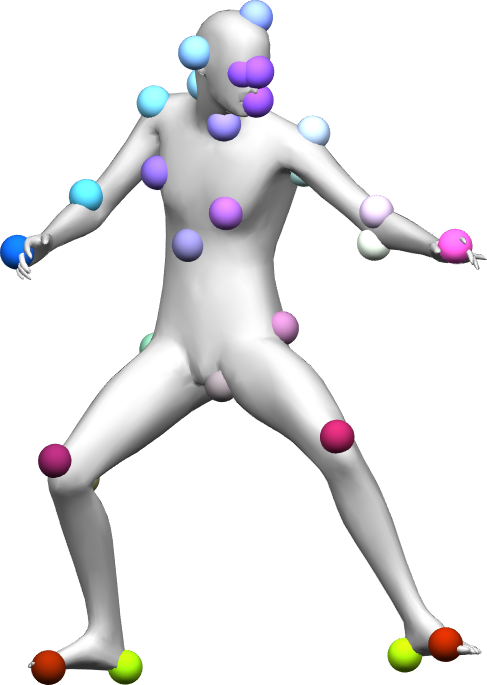}\\
    \includegraphics[height=\heightA]{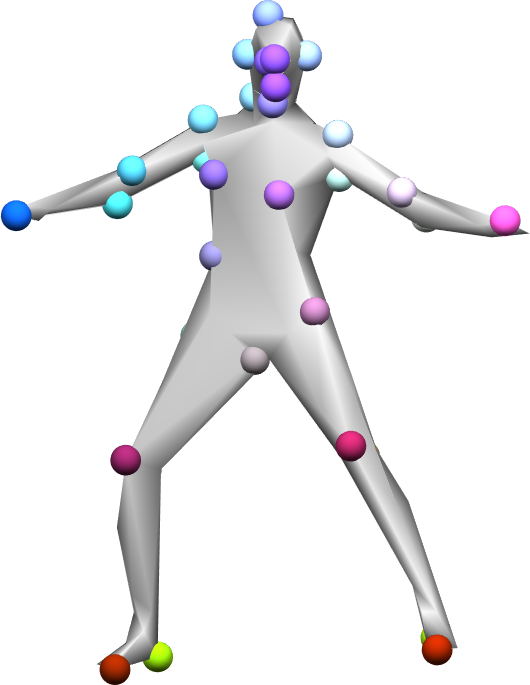}\\
    }
    \tabfig{
    \includegraphics[height=\heightA]{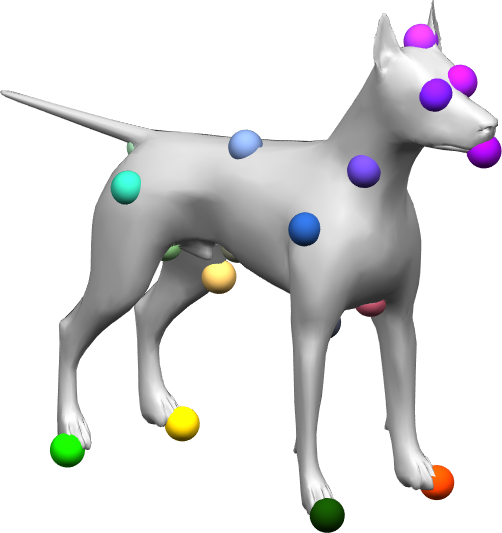}\\
    \includegraphics[height=\heightB]{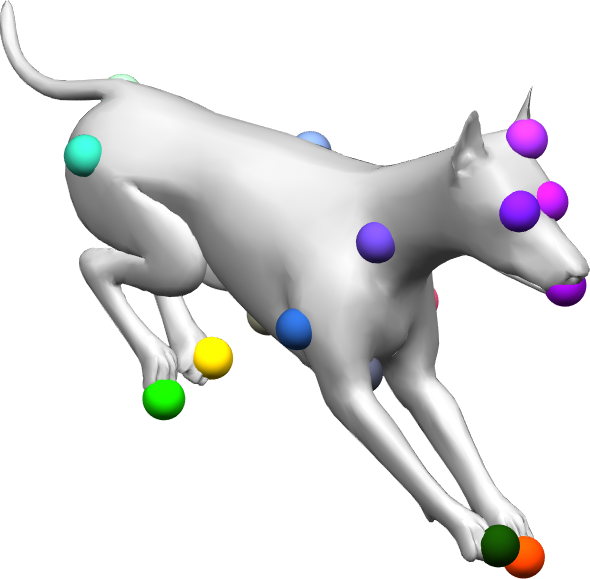}\\
    \includegraphics[height=\heightA]{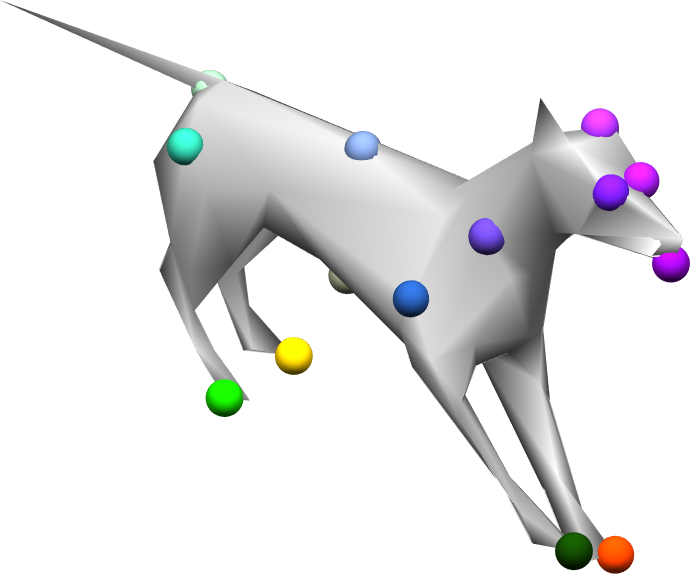}\\
    }
    \tabfig{
    \includegraphics[height=\heightA]{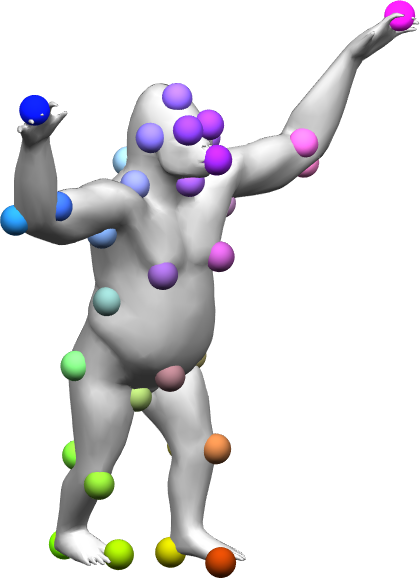}\\
    \includegraphics[height=\heightB]{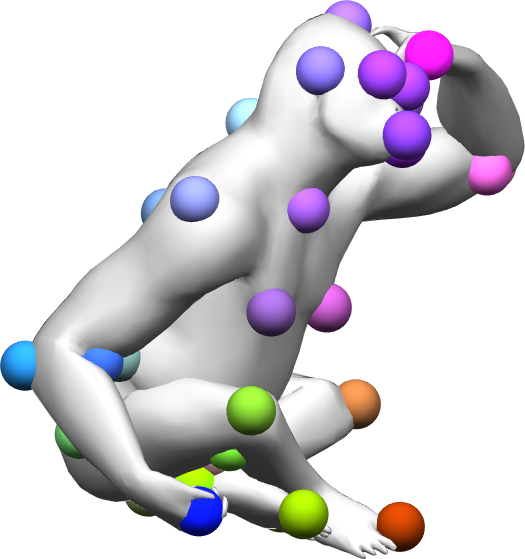}\\
    \includegraphics[height=\heightA]{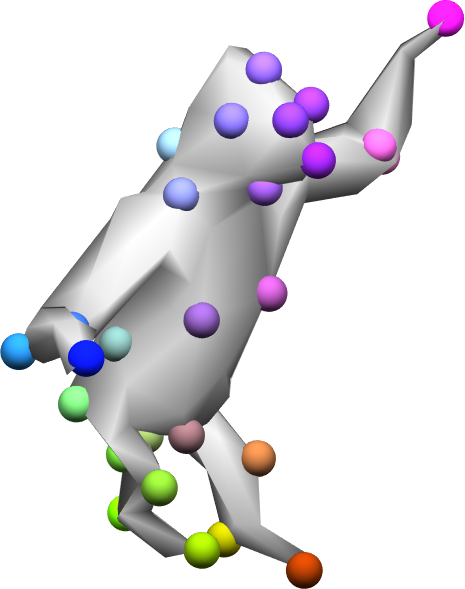}\\
    }
    }
    } 
    \centerline{}
    \centerline{
    \\
    \rule{\linewidth}{0.3pt}
    \\
    }
    \centerline{} 
\centerline{
\noindent
\scalebox{0.67}{
    \tabfig{
    \includegraphics[height=\heightA]{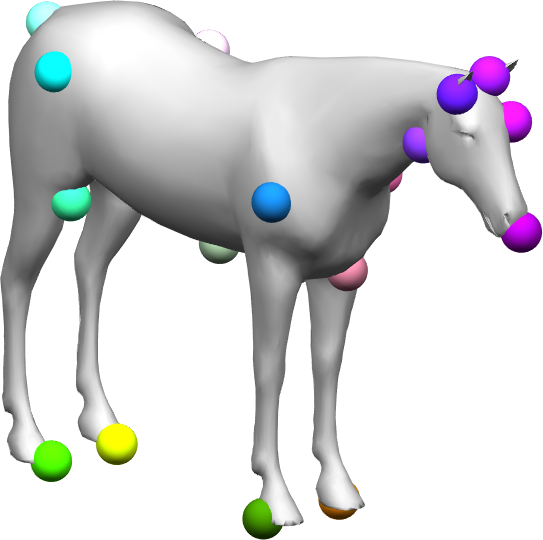}\\
    \includegraphics[height=\heightB]{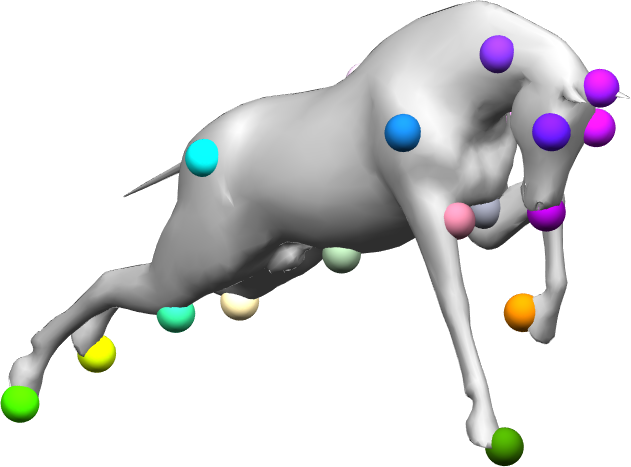}\\
    \includegraphics[height=\heightA]{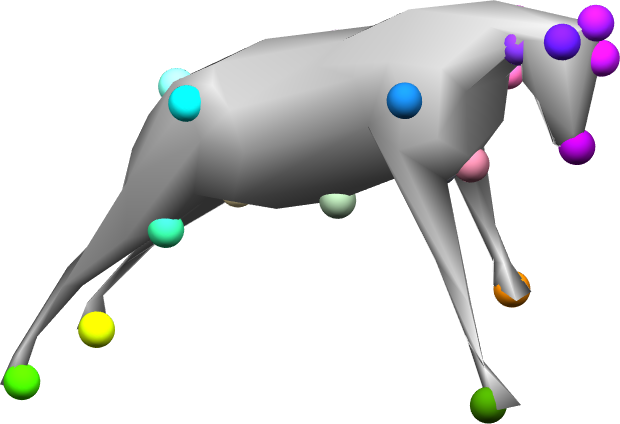}\\
    }
    \tabfig{
    \includegraphics[height=\heightA]{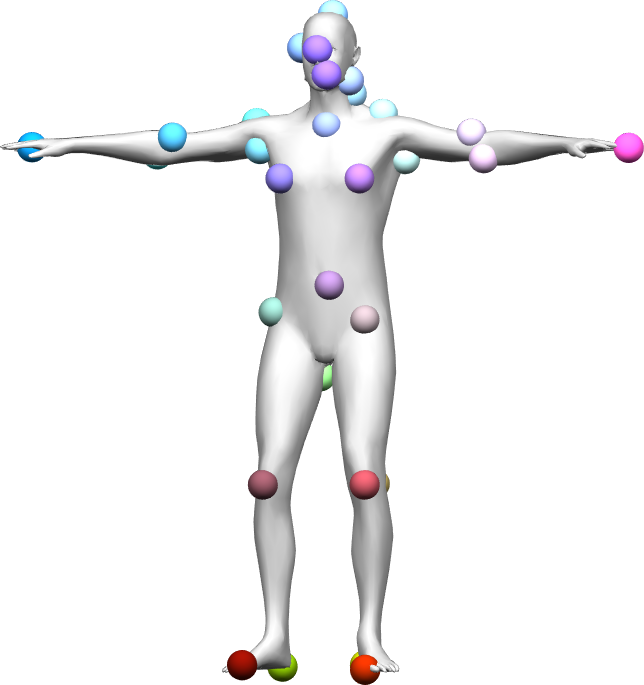}\\
    \includegraphics[height=\heightB]{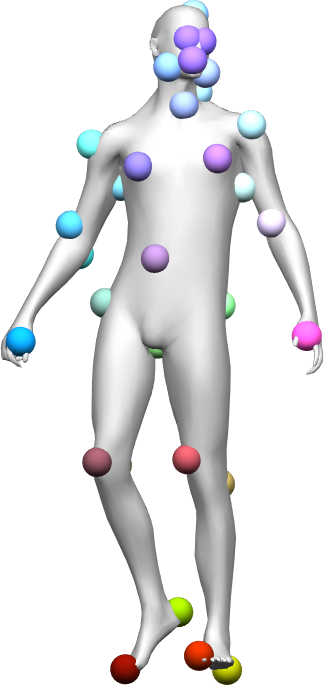}\\
    \includegraphics[height=\heightA]{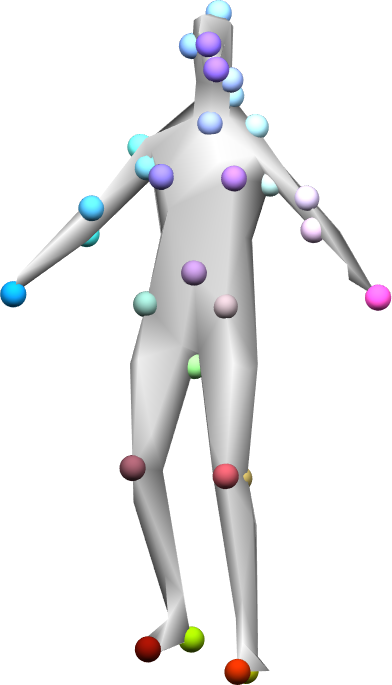}\\
    }
    \tabfig{
    \includegraphics[height=\heightA]{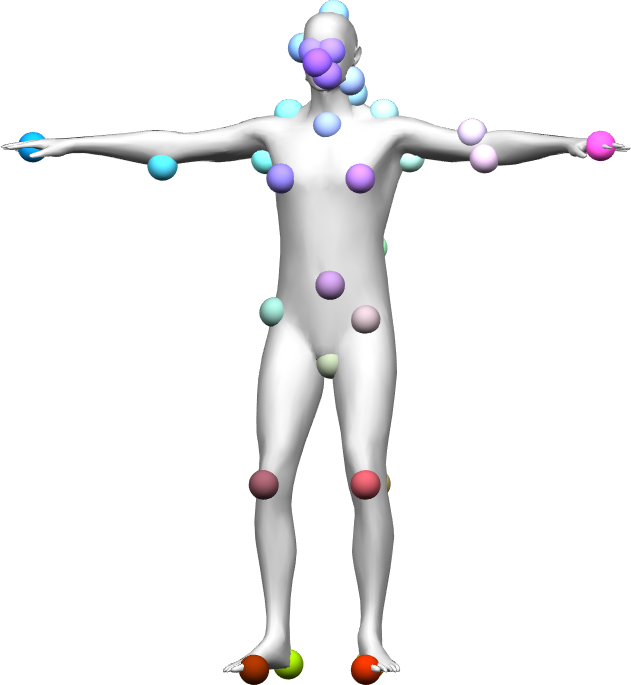}\\
    \includegraphics[height=\heightB]{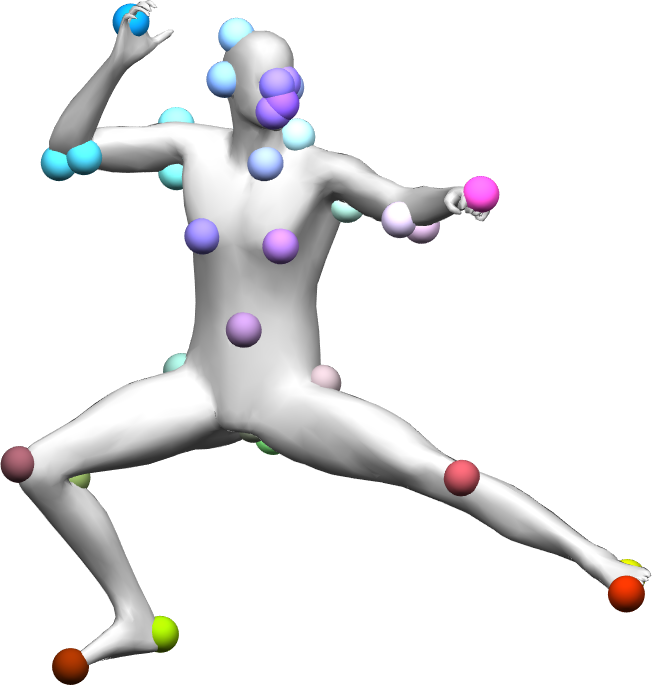}\\
    \includegraphics[height=\heightA]{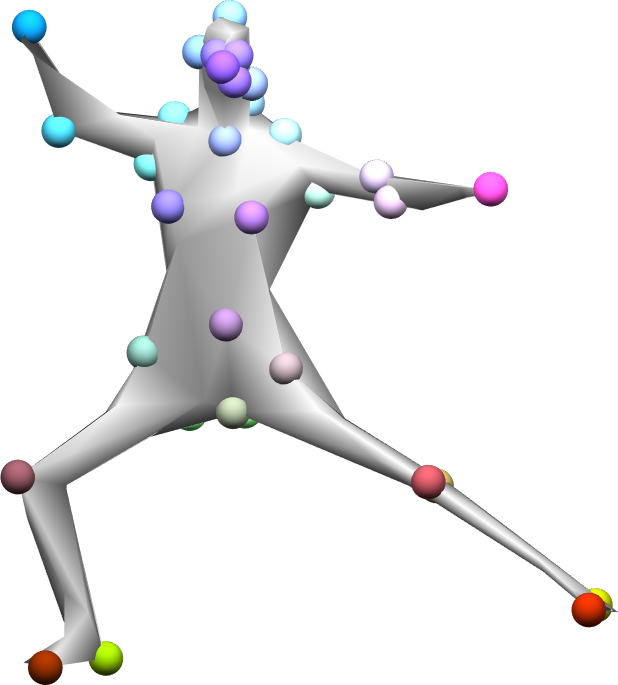}\\
    }   
    \tabfig{
    \includegraphics[height=\heightA]{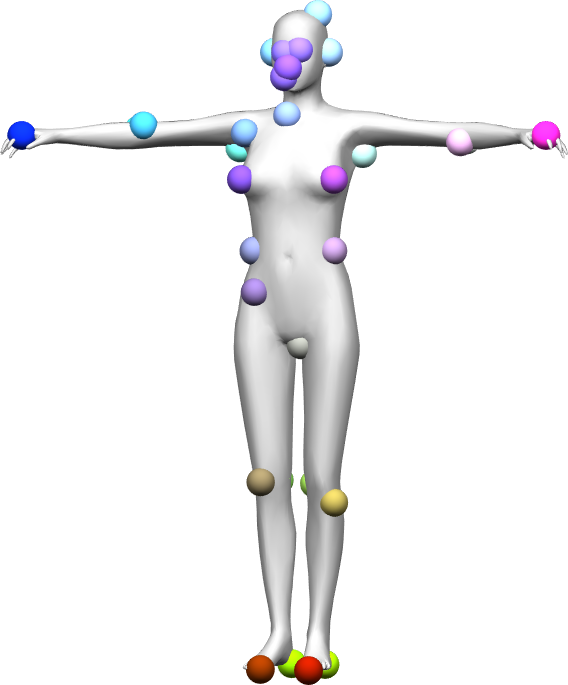}\\
    \includegraphics[height=\heightB]{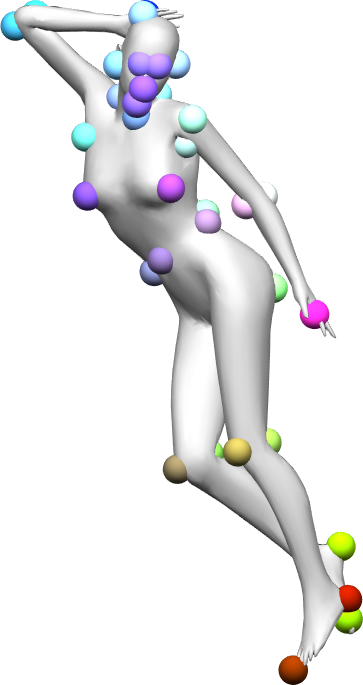}\\
    \includegraphics[height=\heightA]{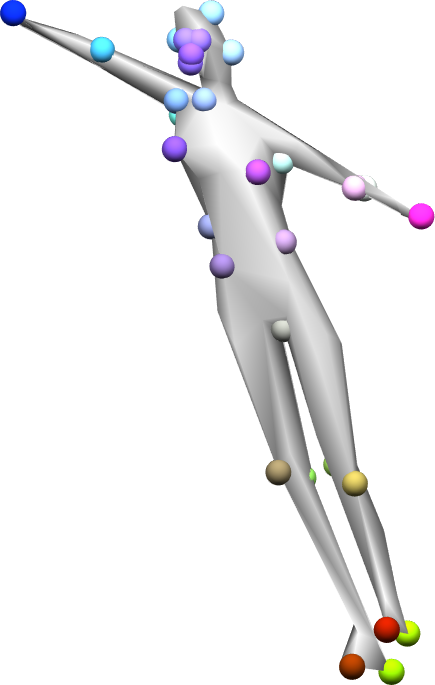}\\
    }  
     \tabfig{
    \includegraphics[height=\heightA]{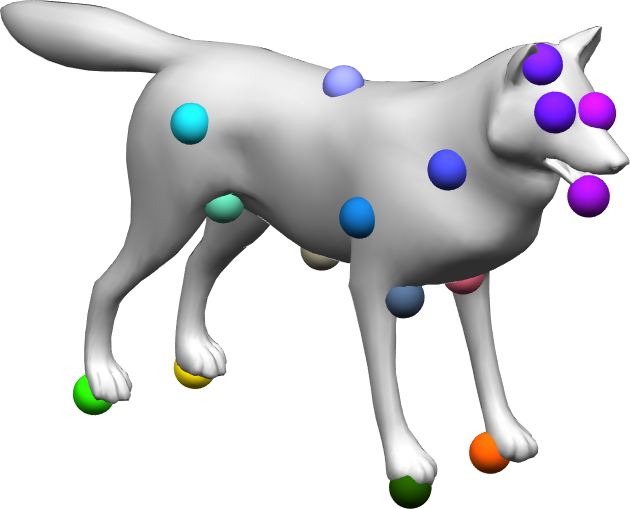}\\
    \includegraphics[height=\heightB]{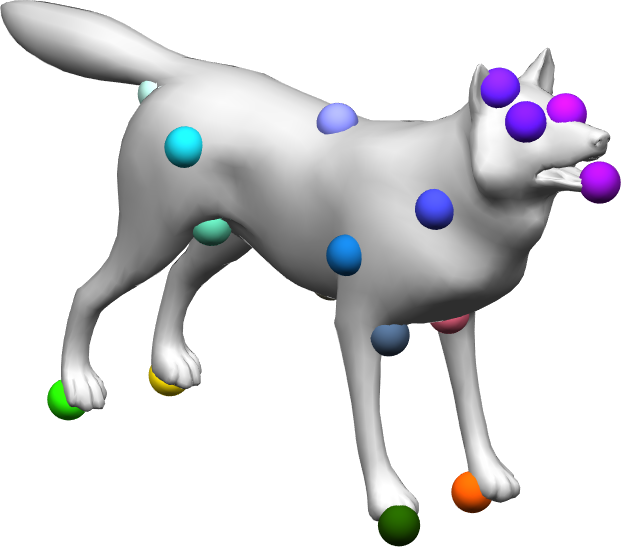}\\
    \includegraphics[height=\heightA]{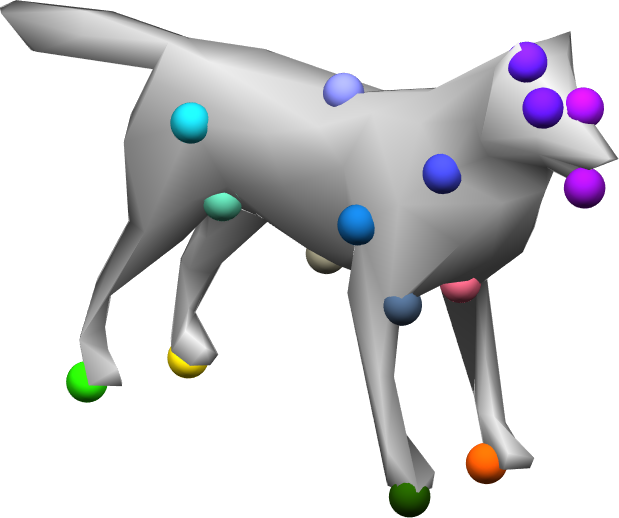}\\
    }
  }
}
    \caption{Correspondences obtained from our method for several shape matching instances from the TOSCA dataset~\cite{Bronstein:2008:NGN:1462123}. Correspondences are indicated by dots with corresponding colours. In each triplet of rows we show $\shapeX$, $\shapeY$, and the deformed shape $\tau(\shapeX)$ from top to bottom. Note that the deformation $\tau$ is not always able to obtain a good alignment (particularly for severe non-rigid transformations,~e.g.~the cat or the gorilla), but the correspondences are still reasonable in many cases.}
    \label{fig:qualitativeResultsToscaSupp} 
\end{figure*}

\end{document}